\documentclass[10pt,twocolumn,letterpaper]{article}

\usepackage{cvpr}              

\newcommand{\model}{MoFlow\xspace}

\definecolor{cvprblue}{rgb}{0.21,0.49,0.74}
\usepackage[pagebackref,breaklinks,colorlinks,allcolors=cvprblue]{hyperref}
\usepackage{amsmath}
\usepackage{amssymb}
\usepackage{mathtools}
\usepackage{amsthm}
\usepackage{thmtools, thm-restate}
\usepackage{wrapfig}
\usepackage{algorithm}
\usepackage{algpseudocode}
\usepackage[accsupp]{axessibility}  

\usepackage{amsmath,amsfonts,bm}

\def\eqref#1{equation~\ref{#1}}

\def\1{\bm{1}}

\DeclareMathAlphabet{\mathsfit}{\encodingdefault}{\sfdefault}{m}{sl}
\SetMathAlphabet{\mathsfit}{bold}{\encodingdefault}{\sfdefault}{bx}{n}

\def\gL{{\mathcal{L}}}

\DeclareMathOperator*{\argmin}{arg\,min}

\usepackage{graphicx}
\usepackage{booktabs} 
\usepackage{multirow}
\usepackage{multicol}
\usepackage{xspace}
\usepackage{xcolor}
\usepackage{transparent}
\usepackage{pifont}

\usepackage{caption}
\usepackage{subcaption}
\usepackage{tikz}

\numberwithin{equation}{section}

\def\paperID{6044} 
\def\confName{CVPR}
\def\confYear{2025}

\title{MoFlow: One-Step Flow Matching for Human Trajectory Forecasting via Implicit Maximum Likelihood Estimation based Distillation}

\author{Yuxiang Fu$^{1,2,*}$, Qi Yan$^{1,2,}$\thanks{Equal Contribution}, Lele Wang$^{1}$, Ke Li$^{4}$, Renjie Liao$^{1,2,3}$\\
$^{1}$University of British Columbia, $^{2}$Vector Institute for AI, \\
$^{3}$Canada CIFAR AI Chair, $^{4}$Simon Fraser University \\
{\tt\small \{yuxiang.fu, qi.yan, lelewang\}@ece.ubc.ca, keli@sfu.ca, rjliao@ece.ubc.ca} \\
}

\begin{document}
\newcommand{\MyIndent}{\hspace{20pt}}
\renewcommand{\b}[1]{\mathbf{#1}}
\renewcommand{\b}[1]{\boldsymbol{#1}}
\renewcommand{\u}[1]{\underline{#1}}
\maketitle
\begin{abstract}
In this paper, we address the problem of human trajectory forecasting, which aims to predict the inherently multi-modal future movements of humans based on their past trajectories and other contextual cues.
We propose a novel motion prediction conditional flow matching model, termed MoFlow, to predict K-shot future trajectories for all agents in a given scene.
We design a novel flow matching loss function that not only ensures at least one of the K sets of future trajectories is accurate but also encourages all K sets of future trajectories to be diverse and plausible.
Furthermore, by leveraging the implicit maximum likelihood estimation (IMLE), we propose a novel distillation method for flow models that only requires samples from the teacher model. 
Extensive experiments on the real-world datasets, including SportVU NBA games, ETH-UCY, and SDD, demonstrate that both our teacher flow model and the IMLE-distilled student model achieve state-of-the-art performance. These models can generate diverse trajectories that are physically and socially plausible.
Moreover, our one-step student model is \textbf{100} times faster than the teacher flow model during sampling. The code, model, and data are available at our project page: \url{https://moflow-imle.github.io/}. 
\end{abstract}    
\section{Introduction}
\label{sec:intro}

Human trajectory forecasting comprises a dual-purpose mission: 1) to predict future trajectories of human with unparalleled precision and 2) to generate diverse future movements that reflect the uncertainty of the complex decision-making processes inherent to human behaviour. 
Pedestrian trajectory prediction task plays an important role in modern autonomous driving systems~\cite{jain2020discrete, luo2021safety, nayakanti2023wayformer, ettinger2021large}, autonomous drones~\cite{floreano2015science} and human-robot interactions~\cite{zhang2020recurrent}.
In a multi-agent scene, trajectory forecasting involves predicting the distribution of future paths through the exploitation of individual historic trajectories and mutual influence among agents. 
Despite significant advancements in recent years~\cite{Gu_2022_CVPR, groupnet_CVPR, xu2024dynamic,Lee_2022_CVPR, liu2021social, bae2024singulartrajectory}, predicting future human trajectories remains a challenging task due to the intrinsic multi-modality of human motion.

\begin{figure}[t]
  \includegraphics[width=\linewidth]{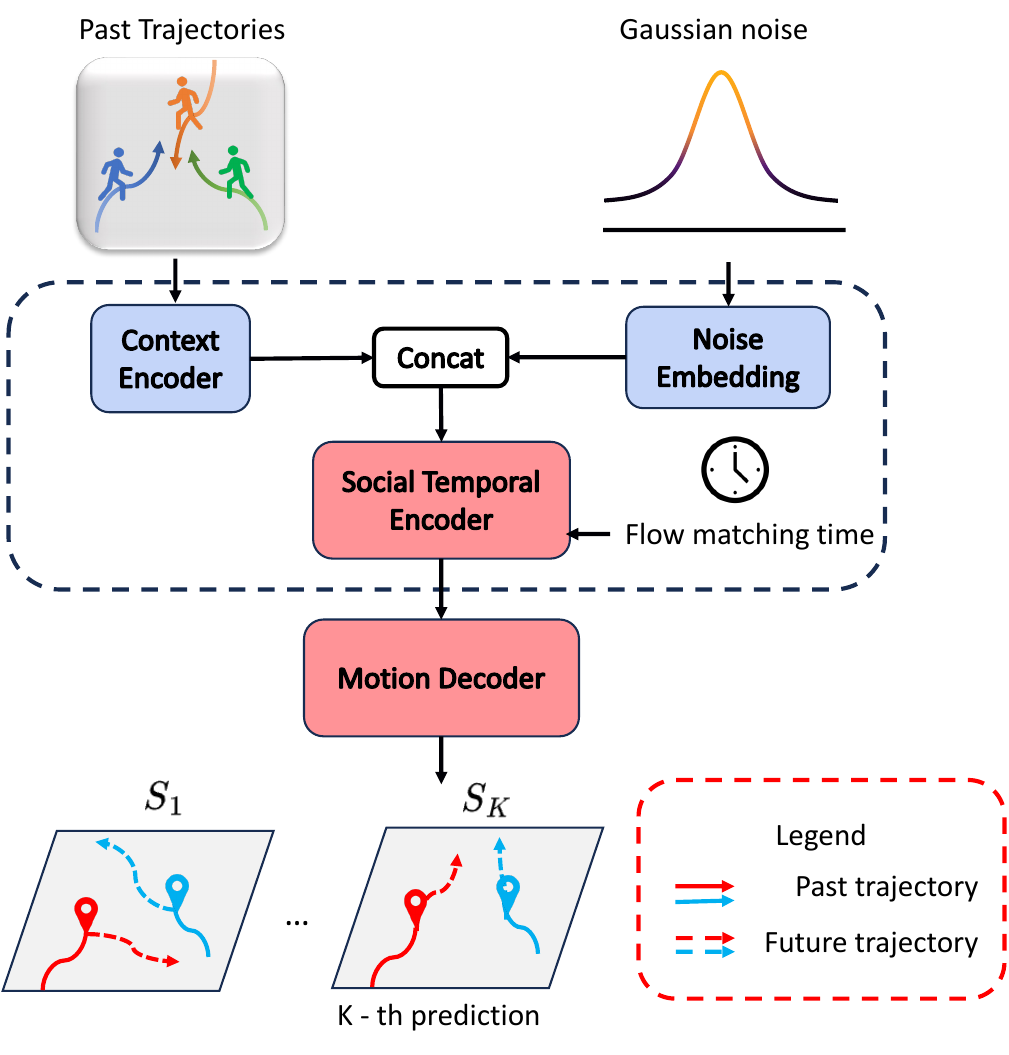}
  \vspace{-0.8cm}
  \caption{
  Overview of our \model model, which consists of a social temporal encoder for contextual cues and a motion decoder that predicts $K$-shot future trajectories of all agents in a scene.}
  \label{fig:intro}
  \vspace{-10pt}
\end{figure}

To tackle this challenge, the majority of previous studies have turned to generative models~\cite{dendorfer2021mg,groupnet_CVPR,xu2024dynamic,Lee_2022_CVPR}, capturing the multi-modal nature of potential future paths via a latent variable. 
Recent works have leveraged the denoising diffusion models (DMs)~\cite{Gu_2022_CVPR, Mao_2023_CVPR, kollovieh2023predict} for probabilistic time series forecasting through their integration with sequence-to-sequence models, due to its outstanding capacity to represent the data distribution. 
Note that DMs can be seen as special cases of flow-based models with conditional Gaussian probability path~\cite{albergo2023stochastic,kingma2018glow, liu2022flow, lipman2022flow}. 
Therefore, we use the term DMs and flow models interchangeably unless specified otherwise.
Motion Indeterminacy Diffusion (MID)~\cite{Gu_2022_CVPR} marks a significant milestone by first applying DMs to human trajectory prediction, where future positions are considered as particles in theromodynamics.
Nonetheless, such diffusion based approach encounters several critical drawbacks. 
First, for each agent, multiple future trajectories from DMs are sampled independently. As a result, it is common for multiple sampled trajectories to overlap, leading to a lack of spatial diversity. This is particularly unsatisfactory when modeling human trajectories, where the inherent multi-modality suggests that future trajectories should often diverge toward different directions and regions.
Second, the sampling process of DMs is notoriously slow and inherently sequential. DMs require a large number of neural function evaluations (NFEs)~\cite{song2023consistency} to numerically integrate a differential equation. 
This limits their applicability to time-critical tasks, such as autonomous driving, and may compromise safety.


In this paper, we aim to address the above challenges with DMs for human trajectory prediction.
The main contributions are outlined as follows: 
\begin{itemize}
    \item We present a novel \textbf{Mo}tion prediction \textbf{Flow} matching (\model) model for trajectory prediction tasks.    
    For each agent in a scene, instead of modeling a single future trajectory, we model multiple future trajectories jointly using the flow model.
    We design a new flow matching loss in tandem that promotes the learning of a diverse set of future trajectories that well capture the multi-modality of human trajectories.
    \item We propose a distillation method for flow models based on Implicit Maximum Likelihood Estimation (IMLE) \cite{li2018implicit}. 
    Compared to existing distillation methods like consistency distillation (CD) \cite{song2023consistency}, it only requires samples from the teacher model, thus being efficient.  
    \item Both \model and the proposed distillation method attain state-of-the-art performance in various metrics on the three human motion datasets, effectively balancing the trade-off between the accuracy and diversity in trajectory forecasting. Our results have the implications as i) our approach accelerates the standard conditional flow matching inference by a great margin, and ii) our IMLE model distills the teacher model in a principled manner that models prediction uncertainty in a sensible way.
\end{itemize}


\begin{figure*}[t]
  \centering
    \includegraphics[width=\linewidth]{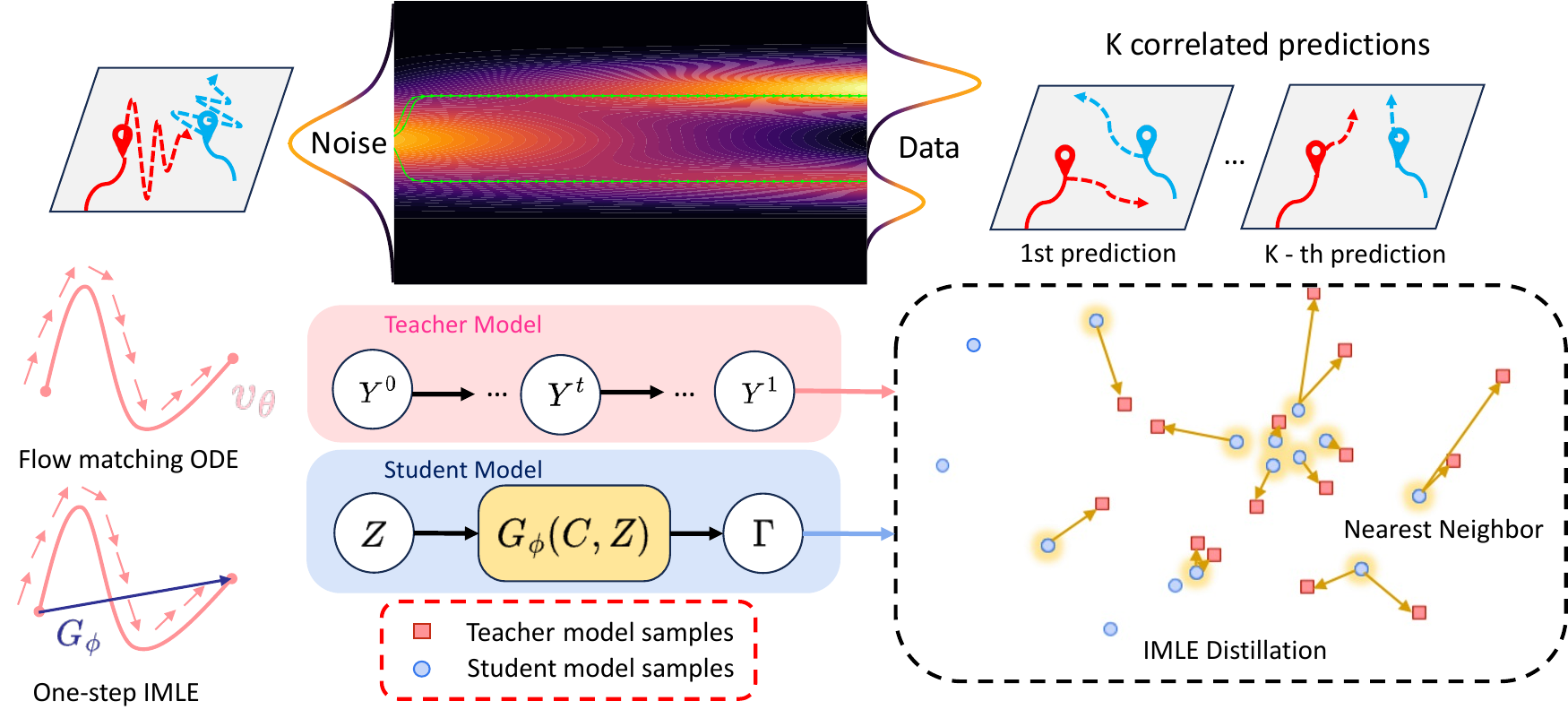}
    \vspace{-0.75cm}
    \caption{
    The overall scheme of our proposed \model model and the IMLE distillation framework.
    \textbf{Top:} Solid lines mean past trajectories; dotted lines show multi-modal future predictions.
    The teacher \model model predicts by solving the denoising ODE, with the green paths being ODE solutions mapping samples from noise to data.
    \textbf{Bottom:} The IMLE objective trains a student model for one-step inference by minimizing the distance between a teacher model sample and its closest counterpart from the student model, as indicated by the arrows.  
    }
  \label{fig:schematic_diagram}
  \vspace{-0.5cm}
\end{figure*}

\section{Related work}
\subsection{Human Trajectory Prediction} 
Human motion trajectories are inherently uncertain~\cite{wong2024socialcircle, Gu_2022_CVPR}, as individuals alter their future motion at will or adjust their direction in response to their surroundings. With a historical record of observed trajectories, numerous viable paths emerge for how pedestrians might move forward. In pursuit of modeling the indeterminacy of human trajectory, recent efforts address the dynamic trajectory prediction problem by using an ensemble of deep generative models. For instance, Generative Adversarial Network (GAN) frameworks~
\cite{gupta2018social, sadeghian2019sophie, dendorfer2021mg} are adapted to generate future stochastic trajectories. Previous works~\cite{groupnet_CVPR,xu2024dynamic,Lee_2022_CVPR} exploit the conditional variational autoencoder (CVAE) to learn distributions via variational inference, while CF-VAE~\cite{bhattacharyya2019conditional} adopts conditional normalizing flow to learn a representative prior.
More recently, DMs have been introduced for human trajectory forecasting tasks, exhibiting remarkable performance and capturing future multi-modality~\cite{Gu_2022_CVPR, Mao_2023_CVPR}.

\subsection{Distillation of Diffusion and Flow Models}
Denoising DMs~\cite{sohl-dickstein15,ho2020denoising,song2019nips} and their extended family of flow-matching methods~\cite{lipman2022flow, liu2022flow, albergo2023stochastic} represent an innovative class of deep generative models inspired by non-equilibrium thermodynamics. The concept behind DMs involves gradually corrupting the data with noise through a forward diffusion process and employing a learned reverse process to reconstruct the original data structure. These models have attracted substantial attention for their state-of-the-art performance across a wide array of generative tasks, such as 3D objects~\cite{zeng2022lion,Wu_2023_CVPR, wu2024consistent3d}, video~\cite{bar2024lumiere, ho2022video, polyak2024movie, melnik2024videodiffusion} and discrete structure graphs~\cite{yan2023swingnn, abramson2024accurate, gat2025discrete}. However, these models suffer from hundreds and thousands of iterative denoising steps required during inference. To accelerate the inference process, many previous works~\cite{song2020denoising, wu2023fast, lu2023dpmsolverfastsolverguided} concentrate on developing faster numerical solvers. Notice that sampling process in DMs fundamentally involves solving ordinary differential equations (ODEs) or stochastic differential equations (SDEs)~\cite{song2021scorebased}, albeit, these approaches still require more than 10 sampling steps to account for discretization error. Distillation of diffusion and flow models emerges with the goal to substitute all the sampling steps with one NFE. In particular, it utilizes a student model to distill the multi-step outputs from the teacher DM~\cite{PD, berthelot2023tract, Meng_2023_CVPR, zheng2023fastsampling, gu2023boot, song2023consistency}. Diffusion distillation~\cite{luo2023comprehensive} incorporates two primary classes: 

(1) \textit{Sampling trajectory dynamics refinement.} Prior works often train a student model to approximate the deterministic process of teacher DM during inference. For instance, knowledge distillation~\cite{luhman2021knowledge, hinton2015distillingknowledgeneuralnetwork} trains a student model to transform noise to data. Progressive distillation (PD)~\cite{PD, Meng_2023_CVPR} involves training a sequence of student models w.r.t the original DM, each requiring half the sampling steps of its predecessor. Rectified flow~\cite{liu2022flow, liu2023instaflow, kim2025simple} simulates denoising ODEs to create synthetic datasets while the ReFlow operation enforces a nearly straight flow to the target distribution. Moreover, CD~\cite{song2023consistency} and TRACT~\cite{berthelot2023tract} train a student model to align its output at one timestep with its output at another timestep along the probability ODE flow, ensuring consistency across multiple timesteps and thus consistent with the teacher. Shortcut models~\cite{frans2025one} unify self-consistency and flow-matching objectives by incorporating both the current noise level and step size, enabling end-to-end training. This distillation technique accelerates convergence by dividing the process into $\log_2(T)$ stages, where the step size is progressively increased, effectively reducing the required bootstrap paths. 

(2) \textit{Distribution alignment.}  The student model distills knowledge by matching its output distribution with that of the teacher diffusion model, leveraging adversarial \cite{sauer2024adversarial} and distribution-matching \cite{yin2025improved} objectives.
Denoising Diffusion GAN~\cite{xiao2022DDGAN} uses GANs for expressive denoising, while Adversarial Distillation Diffusion~\cite{sauer2024adversarial} extends this to the SD-XL Turbo model for efficient image generation. Our method falls into this category. Comparing with other maximum likelihood-based approach like EM distillation~\cite{Xie2024EMDF}, our approach does not incorporate expensive MCMC procedures. There is another array of methods that minimize different divergence measures through variational score distillation~\cite{wang2023prolificdreamer}, including reverse KL divergence~\cite{yin2025improved, yin2024one}, forward KL divergence~\cite{luo2023diff, Xie2024EMDF}, and fisher divergence~\cite{zhou2024score}.

Regarding accelerating a DM that generates human motion trajectory, LEapfrog Diffusion (LED)~\cite{Mao_2023_CVPR} designs a deterministic initializer to predict trajectories corresponding to the latent diffusion state at diffusion time step $\tau$ where $\tau\ll T$ and $T$ is the total diffusion time steps. 
Although this distillation approach enables fast inference, it is often associated with a discernible deterioration in sample quality.

\section{Background}\label{sec:background}

\subsection{Motion Prediction}
The goal of motion prediction is to generate plausible future trajectories for agents of interest based on their motion history and that of surrounding agents. 
Let $\b x = [x_{-T_p+1}, x_{-T_p+2}, \ldots, x_0] \in \mathbb{R}^{2T_p}$ represent the observed past movements over $T_p$ timestamps, where $x_\tau \in \mathbb{R}^2$ captures the spatial coordinates at time $\tau$ from a bird's-eye view. 
Given $A$ agents in the scene, their past trajectories are represented by $X = [\b x_1^{\top}; \b x_2^{\top}; \dots; \b x_A^{\top}] \in \mathbb{R}^{A \times 2T_p}$, where $\b x_i$ denotes the past trajectory of the $i$-th agent and $[\cdot;\cdot]$ means stacking row vectors. 
For simplicity, we use $C \in \mathbb{R}^{A \times C_a T_p}$ to represent all the contextual information, where \( C_a \) is the number of features per timestamp (\eg, absolute coordinates, relative coordinates with respect to the last frame, and velocity).
Let $\b y_i = [y_1, y_2, \ldots, y_{T_f}] \in \mathbb{R}^{2T_f}$ be the ground-truth future trajectory of the $i$-th agent across the future $T_f$ frames, where $y_\tau \in \mathbb{R}^2$ is the coordinate at future time $\tau$.
We denote the future trajectories of $A$ agents in the scene as $Y = [\b y_1^{\top}; \b y_2^{\top}; \dots; \b y_A^{\top}] \in \mathbb{R}^{A \times 2T_f}$.
In principle, the objective of motion prediction is to train a probabilistic model $p_\phi( Y \vert C)$ with parameters $\phi$, which captures the multi-modality and the uncertainty of multi-agent motion dynamics. 
Typically, this model is trained via the maximum likelihood objective
$
\max_{\phi} \mathbb{E}_{(Y, C) \sim p_{\mathcal{D}}} \left[\log p_\phi(Y \vert C) \right],
$
where $p_{\mathcal{D}}$ stands for the data distribution consisting of the coupled context information and future motion trajectories.

\subsection{Flow Matching and Diffusion}
Recent advancements in generative models, including flow-matching models~\cite{lipman2022flow, liu2022flow, albergo2023stochastic} and DMs~\cite{sohl-dickstein15, ho2020denoising, song2020denoising, song2021scorebased}, rely heavily on the effective learning of ODEs or SDEs to transform noise into data.
While differing in specific formulations, diffusion and flow models share similar generative modeling principles~\cite{lipman2022flow, albergo2023stochastic}.
In this paper, we adopt the simple linear flow objective used in previous works~\cite{lipman2022flow, liu2022flow, albergo2023stochastic}. 
Given the observed future scene-level trajectories of $Y^1 \sim q(\cdot)$, we can linearly interpolate between $Y^1\in \mathbb{R}^{A \times 2T_f}$ and a noise variable with the same dimension, $Y^0 \sim \mathcal{N}(\b 0, \b I)$, to obtain $Y^t$, where the superscript \( t \in [0, 1) \) denotes the time in flow models, rather than the time in motion prediction.
In particular, we have, 
\begingroup
\setlength{\abovedisplayskip}{4pt}
\setlength{\belowdisplayskip}{4pt}
\begin{align}\label{eq:flow_interpolation}
    Y^t &= (1-t)\,Y^0 + t\,Y^1, \\
    U^t &= Y^1 - Y^0,
\end{align}
\endgroup
where the velocity field $U^t$ induces a probability path that interpolates from the noise distribution to the data distribution.
Although $U^t$ is a function of flow time $t$ in general, linear flow has a simple constant velocity.
During training, we sample $Y^0$ and $Y^1$ from a fixed prior distribution and the data distribution, respectively, making the the target vector field $U^t$ fully determined~\cite{lipman2022flow}.
We train a neural network $v_\theta$ to approximate the target $U^t$ by minimizing:
\begingroup
\setlength{\abovedisplayskip}{4pt}
\setlength{\belowdisplayskip}{4pt}
\begin{align}
    \mathcal{L}_{\text{FM}} := \mathbb{E}_{Y^t, Y^1, t} \left[ \Vert {v}_\theta(Y^t, C, t) - U^t \Vert^2_2 \right].
    \label{eq:flow_loss}
\end{align}
\endgroup
Here the learned velocity field $v_\theta$ also depends on the contextual information $C$ in the context of motion prediction.
To generate samples from a flow  model, we first sample noise $Y^0$ from the standard normal distribution and then iteratively update it towards \( t = 1 \) according to the learned velocity field, \ie, $Y^{t+1} = Y^{t} + v_\theta(Y^t, C, t)$, which essentially solves the underlying ODE using the Euler’s method.

\section{Method}
\label{sec:method}


We first present our method for encoding context information, then introduce our flow-based teacher model for motion prediction, and finally describe how to use IMLE to distill this teacher model into a student model for one-step sampling. In \cref{fig:schematic_diagram}, the teacher model is shown at the top, and the distillation process is illustrated at the bottom.


\subsection{Context Encoding}

We start by developing an attention-based context encoder to capture the physical dynamics of agents' past motion. Recall that \( C \in \mathbb{R}^{A \times C_a T_p} \) denotes the historical state information for all \( A \) agents. 
To initialize per-agent features, we apply an MLP layer, followed by a transformer encoder module to process inter-agent dynamics:
\begingroup
\setlength{\abovedisplayskip}{4pt}
\setlength{\belowdisplayskip}{4pt}
\begin{align*}
    H_0 &= \mathrm{MLP}(C), \\
    H_0' &= H_0 + \mathrm{PE}_A, \\
    H_{\text{enc}} &= \mathrm{MHSA}({Q}= H_0' , {K}= H_0' , {V}= H_0'),
\end{align*}
\endgroup
where \(\mathrm{PE}_A\) represents the learnable positional encoding based on agent-specific characteristics, such as differentiating between players and the ball in the NBA sports dataset. 
Here $H_0 \in \mathbb{R}^{A \times d}$.
We use the standard multi-head self-attention module in Transformers~\cite{vaswani2017attention}, \(\mathrm{MHSA}(\cdot, \cdot, \cdot)\), to model interactions between agents within each scene.
The context encoder ultimately outputs agent features as \( H_{\text{enc}} \in \mathbb{R}^{A \times d} \), which are then passed to the motion decoder network, as illustrated by ~\cref{fig:intro}. The architectural detail of both the spatio-temporal encoder and motion decoder can be found in the Supplementary Material.


\subsection{Flow Matching for Motion Prediction}
\label{sec:flow_for_motion}
\paragraph{Multi-modal Learning Objective.}
Ensuring diverse motion patterns is essential for the success of trajectory forecasting models. 
Previously, motion generative models have been used to draw a flexible number of $K$ \iid trajectory samples from a given context \( C \), resulting in multi-modal predictions $\{Y_1, Y_2, \cdots, Y_K\}$\footnote{We use the superscript \( t \) to indicate time in flow models and the subscript \( k \) to denote the \( k \)-th predicted future.
} as demonstrated in~\cite{jiang2023motiondiffuser, Gu_2022_CVPR}.
However, their performance is empirically limited~\cite{Mao_2023_CVPR}, as the \iid predicted trajectories fundamentally cannot account for coherence and consistency among \( K \) predictions, such as reducing self-collision and ensuring diverse mode coverage. Although some deterministic methods can generate correlated  $K$  predictions, research on denoising generative models capable of producing correlated  $K$  ensemble predictions remains limited.
To address this issue, we propose using a single flow-matching model with an efficient closest-prediction objective to promote the multi-modality, eliminating the need for extra models or additional training. In contrast, LED~\cite{Mao_2023_CVPR} has to train an additional initializer to match comparable predictive performance. 

By rearranging the linear flow training objective in~\cref{eq:flow_loss} in the data prediction space, 
we introduce a neural network $D_\theta$ and define the equivalent loss:
\begin{align}
    &D_\theta(Y^t, C, t) \coloneqq Y^t + (1-t){v}_\theta(Y^t, C, t), \\
    &\mathcal{L}_{\text{FM}}=\mathbb{E}_{Y^t, Y^1, t} \left[ \frac{ \Vert {D}_\theta(Y^t, C, t) - Y^1 \Vert^2_2}{(1-t)^2} \right].
    \label{eq:loss_fm_data_pred}
\end{align}
Please refer to the Supplement Material for the derivation details. This formulation allows $D_\theta$ to match the future trajectory $Y^1$ in the data space. Namely, $D_\theta$ carries out one-step data prediction given noisy vector $Y^t$ and context $C$ at time $t$. 
It can be reparameterized to obtain $v_\theta$ for ODE sampling.

Next, we design a novel loss to train the model \( D_\theta \) as well as promoting the multi-modality. 
Note that the output dimensionality of \( D_\theta \) in~\cref{eq:loss_fm_data_pred} still reflects a single prediction. We first apply a standard Transformer structure (as shown in~\cref{fig:intro}) to generate \( K \) correlated predictions, where \( D_\theta \) outputs \( K \) scene-level waypoint predictions, denoted by
\(\{S_i\}_{i=1}^K, S_i \in \mathbb{R}^{A\times 2 T_f}\), alongside corresponding classification logits \(\{\zeta_i\}_{i=1}^K, \zeta_i \in \mathbb{R}\).
For simplicity, we omit the time-dependent coefficient in~\cref{eq:loss_fm_data_pred} and apply a combined regression and classification loss as follows:
\begin{align}
    \bar{\mathcal{L}}_{\text{FM}} &= \mathbb{E}_{Y^t, Y^1, t} \left[\Vert S_{j^*} - Y^1 \Vert_2^2 + \mathrm{CE}(\zeta_{1:K}, j^*)\right], 
    \label{eq:loss_fm_reg_cls}
    \\
    j^* &= \arg \min_j \Vert S_j - Y^1 \Vert_2^2,
\end{align}
where $\mathrm{CE}(\cdot, \cdot)$ denotes the cross-entropy loss. 

\paragraph{Input-Output Dimension Adaptation.}
Now, our flow model handles $K$-shot trajectory predictions. 
Its input includes $K$ sets of scene-level noisy trajectories, \(Y^t_{1:K}\).
Note that \(Y^1_i\) represents the \(i\)-th possible configuration of future trajectories of all agents in the scene, whereas \(Y^t_i\) represents the corresponding noisy version at flow time \(t\).  
To solve the denoising ODEs during sampling, we use the \(K\)-correlated trajectory predictions \(\{S_i\}_{i=1}^K\) returned by \(D_\theta\) to compute the \(K\)-shot vector field:
\[
v_\theta^{(i)} = \frac{S_{i} - Y^t_{i}}{1 - t}, \quad \forall i \in [K].
\]
During training, since only one future motion \(Y^1\) is observed in the dataset for a given context \(C\), obtaining interpolated noisy data \(Y^t_{1:K}\) at arbitrary time \(t\) requires a more nuanced approach.  
To stabilize training, we enforce the noise to be shared among all \(K\) components. Specifically, we first draw a noise vector \(Y^0 \sim \mathcal{N}(\b{0}, \b{I})\) and define \(Y^t_i = (1-t) Y^0 + t Y^1, \forall i \in [K]\), ensuring it is identical for all components.  
Additional analysis and ablations on \model can be found in the Supplementary Material.
As for sampling, we initialize \(Y^0_{i}\) from standard normal distribution for all components to solve the ODEs.

\paragraph{Time Schedule.}
Similar to prior works~\cite{esser2024scaling, gat2025discrete, polyak2024movie}, we find that the time schedule used during training significantly impacts model performance. The vanilla uniform distribution \( t \sim \mathcal{U}[0, 1] \) from~\cite{lipman2022flow, liu2022flow} performs sub-optimally on our tasks. Instead, we use a logit-normal distribution for \( t \) with \(\text{logit}(t) \sim \mathcal{N}(\mu_t, \sigma_t^2)\),\footnote{Here, \(\text{logit}(t) = \ln\frac{t}{1-t}\) is the standard logit function.} to better support training.
In particular, the modified loss in~\cref{eq:loss_fm_reg_cls} tends to overfit when \( t \) approaches 1, resulting in larger errors in vector field estimation at the final sample stage.
Thus, we set \(\mu_t = -0.5\) and \(\sigma_t = 1.5\) to ensure that training more frequently covers the noisier regime when $t$ is closer to 0.

\begin{algorithm}[t] 
    \caption{Distilling \model with IMLE}   
    \label{alg:imle-opt}
        \begin{algorithmic}[1]
        \State\textbf{Require:}\xspace Dataset $p_\mathcal{D}$, IMLE sampling size $m$, learning rate $\eta$, teacher flow model $v_\theta$
        \State Initialize the parameters $\phi$ of the network $G_\phi$.
        \Repeat 
            \State $(\cdot, C) \sim p_\mathcal{D}$ \Comment{Sample context data}
            \State $Y^0_1, \ldots, Y^0_K \sim \mathcal{N}(\b 0, \b I)$
            \State Solve the ODE via $v_\theta$ to obtain $\hat{Y}^1_{1:K}$ 
            \vspace{1pt}
            \State $ Z_1,\ldots, Z_m \sim \mathcal{N}(\b 0, \b I)$
            \State $\Gamma_j \gets G_\phi(C, Z_j)$, $j\in [m]$
            \State ${\pi}\gets\argmin_j\gL_{\text{IMLE}}(\hat{Y}^1_{1:K},  \Gamma_j)$   
            \State $\phi\gets\phi-\eta\nabla_{\phi} \gL_{\text{IMLE}}(\hat{Y}^1_{1:K}, \Gamma_{{\pi}})$
        \Until{converged}
        \end{algorithmic}
\end{algorithm}

\subsection{IMLE Distillation for MoFlow}
In this section, we propose a principled distillation method for the learned \model to bypass time-consuming ODE-based sampling. 
The approach leverages conditional implicit maximum likelihood estimation (IMLE)~\cite{li2018implicit, li2019diverse} to train a student model for one-step sampling. Note that IMLE was originally proposed to train generative models that can alleviate the common mode collapse issue.

The IMLE distillation process is outlined in~\cref{alg:imle-opt}, where we adapt this to our trajectory forecasting context. Specifically, lines 4–6 describe the standard ODE-based sampling of the teacher model, where the trained \model numerically solves the ODE up to $t=1$. This produces $K$ correlated multi-modal trajectory predictions $\hat{Y}^1_{1:K}$ conditioned on the context $C$. A conditional IMLE generator $G_\phi$ then uses a noise vector $Z$ and context $C$ to stochastically generate $K$-component trajectories $\Gamma \in \mathbb{R}^{K \times A \times 2T_f}$, matching the shape of $\hat{Y}^1_{1:K}$.

The conditional IMLE objective generates more samples than those present in the distillation dataset for the same context $C$. In practice, $m$ \iid samples are generated via $G_\phi$ for the same context, and the one closest to the teacher model's prediction $\hat{Y}^1_{1:K}$ is selected for loss computation. The objective minimizes the distance between the teacher model sample and its nearest sample from the student model using the distance metric $\mathcal{L}_{\text{IMLE}}$, ensuring that the teacher model data mode is closely approximated by at least one sample from the student model. 
To preserve trajectory prediction multi-modality, we employ the Chamfer distance~\cite{lin2024infoCD} as the loss function $\mathcal{L}_{\text{IMLE}}$, which measures the similarity between two sets of trajectories across all agents:
\begin{align*}
    \resizebox{\columnwidth}{!}{$
    \mathcal{L}_{\text{IMLE}}(\hat{Y}^1_{1:K}, \Gamma) = \dfrac{1}{K} \left( \sum\limits_{i=1}^K \underset{j}{\min} \|\hat{Y}^1_i - \Gamma^{(j)}\| + \sum\limits_{j=1}^K \underset{i}{\min} \|\hat{Y}^1_i - \Gamma^{(j)}\| \right),
    $}
\end{align*}
where $\Gamma^{(i)} \in \mathbb{R}^{A \times 2T_f}$ is the $i$-th component of the IMLE-generated correlated trajectory.

\paragraph{Efficient and Flexible Distillation.}
The IMLE framework provides an efficient and flexible solution for flow model distillation. 
Its backbone, denoted by \(G_\phi\), is architecturally unrestricted and can share the same architecture as the teacher model (except for eliminating time step embeddings), as both teacher and student models aim to learn spatio-temporal dependencies.
Moreover, IMLE is inherently stable compared to adversarial models like GANs, avoiding mode collapse and vanishing gradients~\cite{li2018implicit, li2019diverse}. 
Compared to existing deterministic methods~\cite{Mao_2023_CVPR}, IMLE employs a principled probabilistic approach~\cite{li2018implicit, li2019diverse}, enhancing robustness and flexibility.

\begin{table*}[h]
\centering
\caption{Comparison with baseline models on \textbf{NBA} dataset. min\textsubscript{20}ADE/min\textsubscript{20}FDE (meters) are reported. Bold/underlined fonts represent the best/second-best result. 
For LED, we denote its stage-one DDPM model as LED$^\dagger$ and the results from the LED initializer as LED*.
}
\label{tab:nba_result}
\vspace{-0.25cm}
\begin{tabular}{c|cccccc|cc}
\toprule
\multirow{3}{*}{Time} & MemoNet&NPSN&GroupNet& MID& LED$^\dagger$ & LED*& \textbf{\model} & \textbf{IMLE}  \\
&~\cite{memonet_CVPR} &~\cite{Bae_2022_CVPR} &~\cite{groupnet_CVPR} &~\cite{Gu_2022_CVPR}&~\cite{Mao_2023_CVPR}&~\cite{Mao_2023_CVPR} & & \\
\midrule
1.0s & 0.38/0.56 & 0.35/0.58& 0.26/0.34 & 0.28/0.37 & \u{0.21}/0.28 & \textbf{0.18}/\u{0.27} & \textbf{0.18}/\textbf{0.25} & \textbf{0.18}/\textbf{0.25} \\
2.0s &  0.71/1.14 &0.68/1.23 & 0.49/0.70 &0.51/0.72& 0.44/0.64 & \u{0.37}/\u{0.56} & \textbf{0.33}/\textbf{0.46} & \textbf{0.34}/\textbf{0.47} \\
3.0s &1.00/1.57& 1.01/1.76 &0.73/1.02 &0.71/0.98 & 0.69/0.95 & \u{0.58}/\u{0.84} & \textbf{0.51}/\textbf{0.67} & \textbf{0.52}/\textbf{0.67} \\
Total (4.0s) & 1.25/1.47 &1.31/1.79 &0.96/1.30 &0.96/1.27 & 0.94/1.21 & \u{0.81}/{1.16} & \textbf{0.70}/\u{0.88} & \textbf{0.71/0.86} \\
\bottomrule
\end{tabular}
\end{table*}

\begin{table*}[ht]
\centering
\caption{Comparison with baseline distillation models on \textbf{ETH-UCY} dataset. min\textsubscript{20}ADE/min\textsubscript{20}FDE (meters) are reported. Bold/underlined fonts represent the best/second-best result. LED* represents the results from student model LED initializer.}
\label{tab:eth_ucy_result}
\vspace{-2pt}
\resizebox{\linewidth}{!}{%
\begin{tabular}{c|ccccccc|cc}
\toprule
\multirow{3}{*}{Subsets} &MID&GroupNet& TUTR& EqMotion& EigenTraj & LED*&SingularTraj&\textbf{\model} & \textbf{IMLE}   \\
&~\cite{Gu_2022_CVPR} &~\cite{groupnet_CVPR} &~\cite{Shi_2023_ICCV} &~\cite{xu2023eqmotion} &~\cite{bae2023eigentrajectory} & ~\cite{Mao_2023_CVPR}&~\cite{bae2024singulartrajectory} & \\
\midrule
ETH &  0.39/0.66 & 0.46/0.73 & 0.40/0.61 &0.40/0.61 &\u{0.36}/\u{0.53} & {0.39}/0.58 & \textbf{0.35}/\textbf{0.42} & 0.40/0.57 & 0.40/0.58 \\ 
HOTEL &   0.13/0.22 & 0.15/0.25 & \textbf{0.11}/\u{0.18} & \u{0.12}/\u{0.18}&\u{0.12}/0.19 & \textbf{0.11}/\textbf{0.17} &0.13/0.19 & \textbf{0.11}/\textbf{0.17} & 0.12/0.18 \\ 
UNIV &  \textbf{0.22}/0.45 & 0.26/0.49 & \u{0.23}/0.42 & \u{0.23}/0.43&0.24/0.34 & 0.26/0.44 & 0.25/0.44 & \u{0.23}/\textbf{0.39} & \u{0.23}/\textbf{0.39} \\ 
ZARA1 &  0.17/0.30 & 0.21/0.39 & 0.18/0.34 &0.18/\u{0.32} & 0.19/0.33 & 0.18/\textbf{0.26} & 0.19/0.32 & \textbf{0.15}/\textbf{0.26} & \u{0.16}/\textbf{0.26} \\ 
ZARA2 &  \u{0.13}/0.27 & 0.17/0.33 & \u{0.13}/0.25 &\u{0.13}/\u{0.23}& {0.14}/0.24 & \u{0.13}/\textbf{0.22} &0.15/0.25 & \textbf{0.12}/\textbf{0.22} & \u{0.13}/\textbf{0.22} \\ 
\midrule
AVG  &  \u{0.21}/0.38 & {0.25}/0.44 & \u{0.21}/0.36 &\u{0.21}/0.35 & \u{0.21}/\u{0.33} & \u{0.21}/\u{0.33} & \u{0.21}/\textbf{0.32} & \textbf{0.20}/\textbf{0.32} & \u{0.21}/\u{0.33}\\ 
\bottomrule
\end{tabular}
}
\vspace{-0.25cm}
\end{table*}

\section{Experiments}
\subsection{Datasets and Metrics}
We evaluate our method on three human motion prediction datasets, including one sports dataset (NBA) and two pedestrian datasets (ETH-UCY and Stanford Drone Dataset).
\paragraph{NBA Dataset.} 
The NBA SportVU dataset tracks the trajectories of 10 players and the ball during NBA basketball games. Each game segment has a fixed number of agents. In this task, we predict 20 future frames (4.0s) using 10 past frames (2.0s) as input. 
\newline
\noindent\textbf{ETH-UCY Datasets.}
The ETH-UCY datasets include 1,536 pedestrians across five distinct scenes: ETH, Hotel, Univ, Zara1, and Zara2. These trajectories, recorded from surveillance viewpoints, are labeled in world coordinates, encompassing a range of motion patterns. We use an 8-second segment length to align with the previous works~\cite{Mao_2023_CVPR,Gu_2022_CVPR}, where 8 frames (3.2s) of  past trajectories are used to forecast future 12 frames (4.8s). For evaluation, we adopt a leave-one-out approach, training on four scenes and using the remaining scene for testing.
\newline
\noindent\textbf{SDD dataset.}
The Stanford Drone Dataset (SDD) is a massive pedestrian dataset captured from a bird’s-eye view on a university campus. Following prior studies~\cite{Mao_2023_CVPR,Gu_2022_CVPR, Shi_2023_ICCV}, we use the standard train-test split, utilizing 8 frames (3.2s) of past observations to predict the subsequent 12 frames (4.8s).

\noindent\textbf{Evaluation Metrics.}
1) Minimum Average Displacement Error (min$_K$ADE) computes the minimum Euclidean average distance overall estimated positions in the $K$ predicted trajectories and the ground-truth.
2) Minimum Final Displacement Error (min$_K$FDE) delineates the minimum Euclidean distance between the ground-truth final destination and the final destination of $K$ predicted trajectories. 

\subsection{Implementation Details} \label{implementation}

In our approach, trajectories are preprocessed using min-max normalization to scale future relative motions to the range $[-1,1]$ to stabilize training dynamics. The backbone of our method is a standard transformer architecture~\cite{vaswani2017attention}, which is augmented with sinusoidal positional encodings to capture inter-agent spatial relationships. These encodings are applied at both the agent and prediction levels, with self-attention alternately operating across these levels to reinforce agent interactions and enhance scene coherence. The teacher model generates samples through a 100-step denoising ODE process, and these precomputed samples are subsequently used to evaluate performance and to train the student model via the IMLE objective. The student model retains the same architectural structure as the teacher, except that it omits the flow time positional encoding layers. 

\subsection{Baselines}
We compare our \model model and IMLE distillation method with state-of-the-art approaches, including MemoNet~\cite{memonet_CVPR}, NPSN~\cite{Bae_2022_CVPR}, GroupNet~\cite{groupnet_CVPR}, MID~\cite{Gu_2022_CVPR}, LED~\cite{Mao_2023_CVPR}, TUTR~\cite{Shi_2023_ICCV}, EqMotion~\cite{xu2023eqmotion}, EigenTraj~\cite{bae2023eigentrajectory}, SingularTraj~\cite{bae2024singulartrajectory}, Evo-Graph~\cite{li2020Evolvegraph}, Y-net~\cite{mangalam2021goals}, CAGN~\cite{CAGN_2022}, and SocialVAE~\cite{socialvae2022}, on datasets where comparable results are readily available.
In terms of LED~\cite{Mao_2023_CVPR}, we reproduce the results using its codebase, while the outcomes for the remaining models are directly extracted from their respective papers.



\begin{figure*}[h]
    \centering
    \begin{subfigure}{0.598\linewidth}
        \centering
        \includegraphics[width=\linewidth]{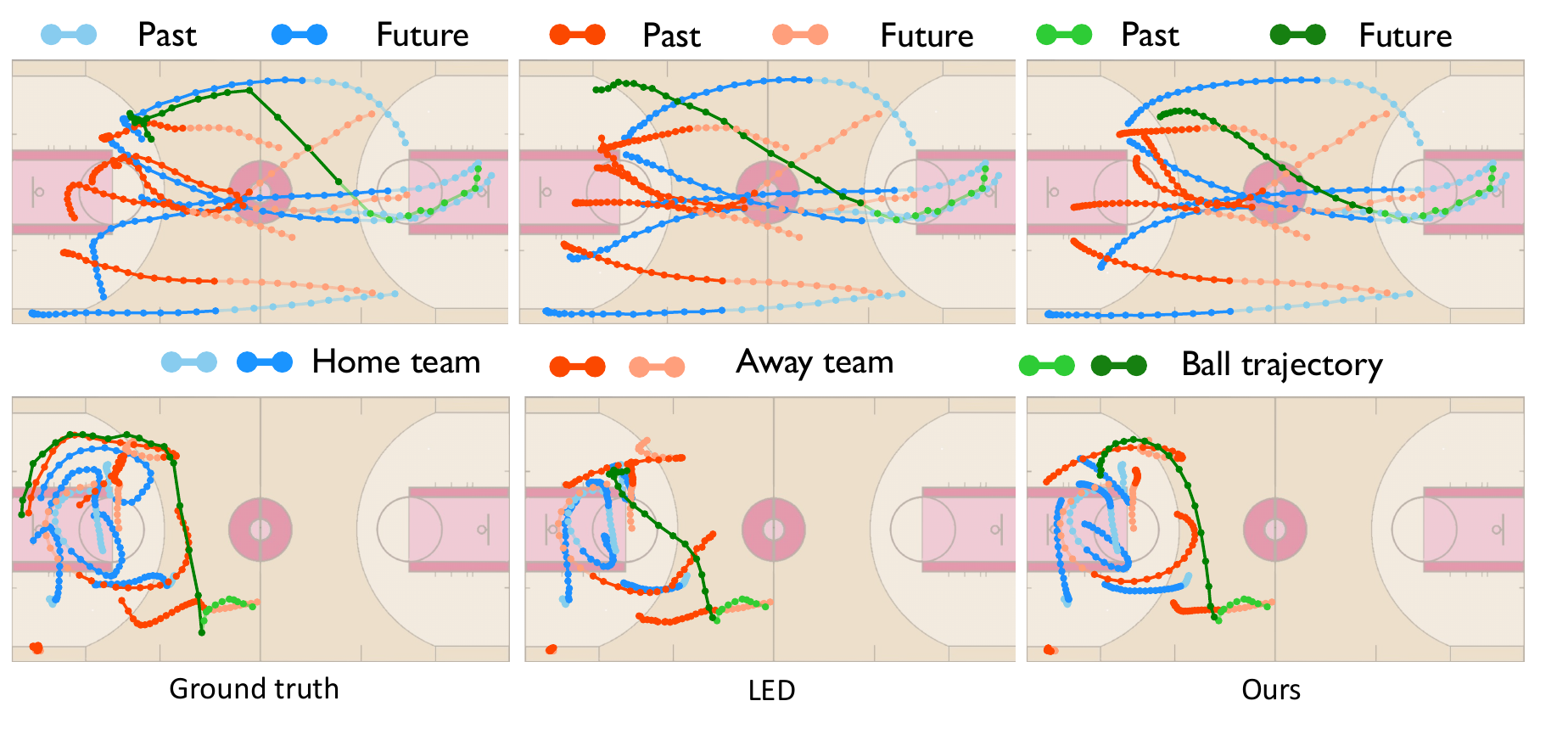}
        \vspace{-0.5cm}
        \label{fig:qualresult_accuracy_nba}
        \caption{}
    \end{subfigure}
    \hfill
    \begin{subfigure}{0.397\linewidth}
        \centering
        \includegraphics[width=\linewidth]{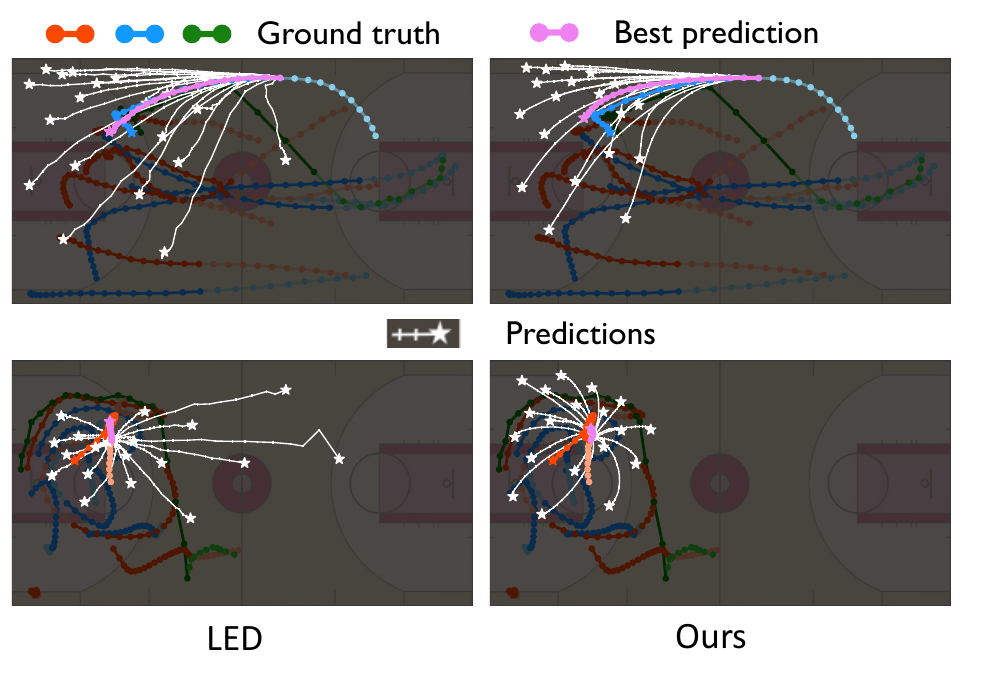}
        \vspace{-0.5cm}
        \label{fig:qualresult_diversity}
        \caption{}
    \end{subfigure}
    \vspace{-0.4cm}
    \caption{Qualitative results on the NBA dataset. (a) We compare between the best-of-20 predictions from our \model IMLE distillation method, the best-of-20 predictions the LED method, and the ground truth future trajectories. The visualization demonstrates that our approach produces predictions that more closely align with the ground truth trajectories compared to the LED model. (b) We are using two same scenes as (a). This figure delineates the diversity of samples from our IMLE generator. Our method generates a prediction that conforms to the GT trajectory. (Pink: the sample closest to the ground truth in $L_2$ sense among $K=20$ predictions.)}
    \label{fig:qualresult_combined}
\end{figure*}

\begin{figure*}[h]
    \centering
    \includegraphics[width=\linewidth]{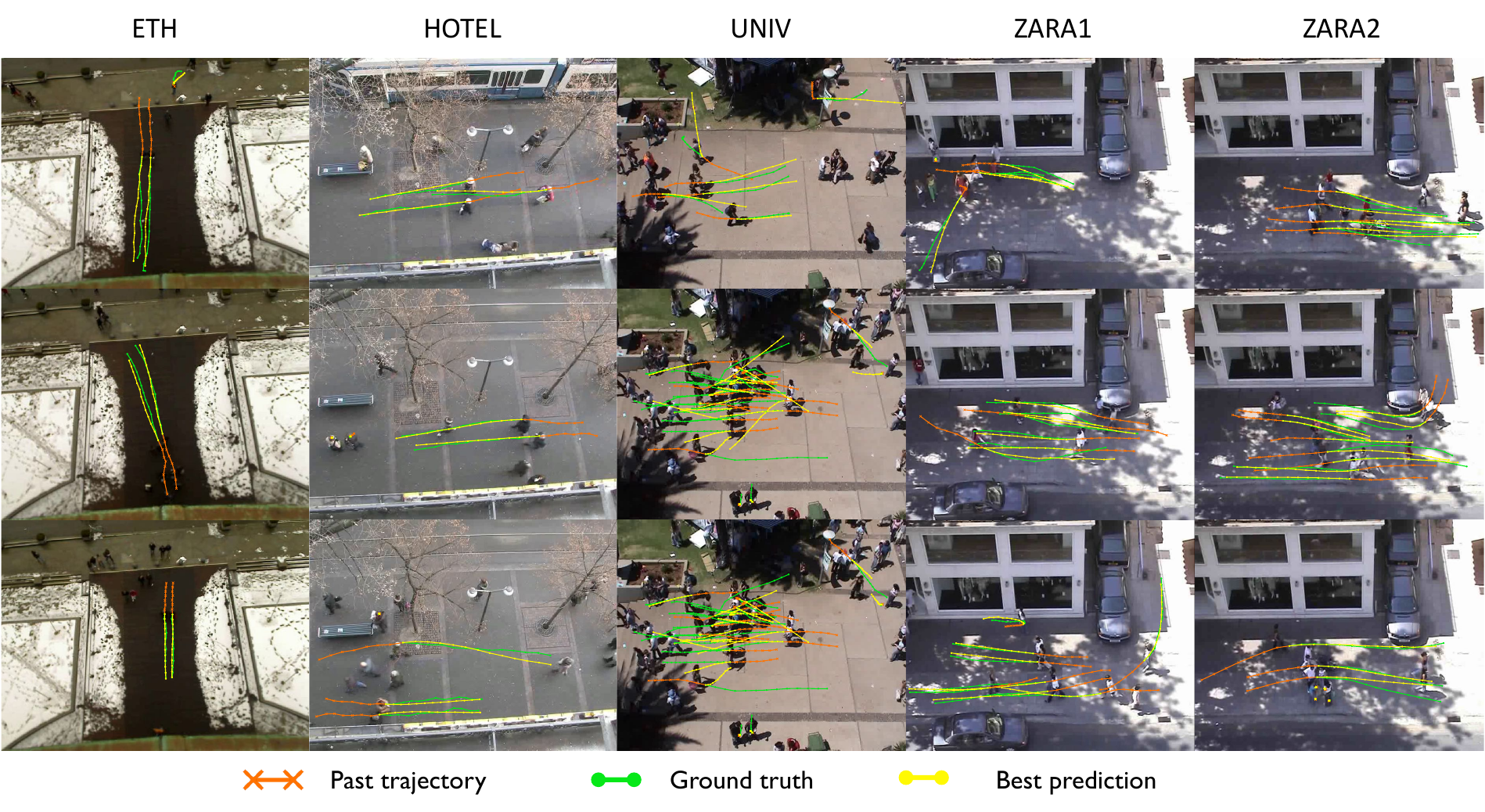}
    \vspace{-0.9cm}
    \caption{
    The qualitative results on the ETH-UCY dataset show that our \model IMLE distillation model’s best-of-20 predictions (selected via the lowest FDE) closely match the ground truth future trajectories, capturing important motion nuances.
    }
    \label{fig:qualresult_accuracy_eth}
\end{figure*}

\subsection{Quantitative Results}
We present our model's state-of-the-art quantitative results in~\cref{tab:nba_result},~\cref{tab:eth_ucy_result}, and~\cref{tab:sdd} across three datasets.  
As shown in~\cref{tab:nba_result}, both our teacher and student methods significantly outperform other baseline models on the NBA dataset, achieving 0.71 minADE and 0.87 minFDE\textemdash improvements of over 12\% and 25\%, respectively, compared to the previous state-of-the-art method.
On the ETH-UCY datasets, our methods deliver performance comparable to the best existing models. It is worth noting that the performance on the ETH-UCY datasets has plateaued in recent years due to the datasets' size and the rapid advancement of expressive models. Nonetheless, our models successfully match the performance of the best available methods.
On the SDD dataset in~\cref{tab:sdd}, our models achieve state-of-the-art performance on the minADE metric and competitive results on the minFDE metric, coming within 0.5 of the best model. 

Notably, with our IMLE distillation, the final performance remains largely unchanged among all three datasets. This is particularly significant, as one-step generation eliminates the need for the time-consuming denoising ODE solving process, which is 100 times faster in our current setup. The retained performance indicates that iterative denoising steps can be effectively removed, enabling faster sampling during deployment. More results about inference time and performance are included 
in the ablation table in the Supplementary Material.

\begin{table}[t]
	\centering
	\caption{Comparison with state-of-the-art methods on the \textbf{SDD} dataset in min\textsubscript{20}ADE/min\textsubscript{20}FDE (pixels).}
        \vspace{-0.25cm}
	\begin{tabular}{l|c|c}
		\toprule
		\textbf{Method} & \textbf{Venue}&\textbf{ADE/FDE} \\ 
		\midrule
		Evo-Graph~\cite{li2020Evolvegraph} & NIPS'20& 13.90/22.90 \\ 
		Y-net~\cite{mangalam2021goals} & ICCV'21& 11.49/20.23 \\ 
		GroupNet~\cite{groupnet_CVPR} & CVPR'22& 9.31/16.11 \\ 
		CAGN~\cite{CAGN_2022} & AAAI'22&9.42/15.93 \\ 
		NPSN~\cite{Bae_2022_CVPR} & CVPR'22&8.56/11.85 \\ 
		MemoNet~\cite{memonet_CVPR} & CVPR'22&9.50/14.78 \\ 
		SocialVAE~\cite{socialvae2022} & ECCV'22&8.88/14.81 \\ 
		MID~\cite{Gu_2022_CVPR} & CVPR'22&9.73/15.32 \\ 
		TUTR~\cite{Shi_2023_ICCV} & ICCV'23&\underline{7.76}/12.69 \\
       EigenTraj~\cite{bae2023eigentrajectory} &
        ICCV'23 & 8.05/13.25 \\
		LED~\cite{Mao_2023_CVPR} &CVPR'23& 8.48/\u{11.66} \\ 
    ET+HighGraph~\cite{Kim_2024_CVPR} &
        CVPR'24 & 7.81/\textbf{11.09} \\
		\midrule
		\textbf{\model} &-&\textbf{7.50}/11.96 \\ 
		\textbf{IMLE} &-&7.85/12.86 \\ 
		\bottomrule
	\end{tabular}
    \vspace{-0.10cm}
    \label{tab:sdd}
\end{table}


\begin{table}[t]
\centering
\caption{
Comparison with different distillation methods on NBA dataset. min\textsubscript{20}ADE/min\textsubscript{20}FDE are reported. 
All the models learn to generate samples conforming to the marginal distribution from teacher model samples at time step $t=1$.}
\label{tab:ablation_nba}
\resizebox{\linewidth}{!}{%
\begin{tabular}{c|ccc}
\toprule
Time & \textbf{\model-IMLE}  & \textbf{WGAN~\cite{arjovsky2017wasserstein}} & \textbf{DCGAN~\cite{radford2016unsupervised}} \\
\midrule
1.0s & 0.18/0.25 & 0.35/0.58 & 0.33/0.60 \\
2.0s & 0.35/0.47 & 0.74/1.56 & 0.72/1.64 \\
3.0s & 0.52/0.67& 1.30/2.30 & 1.22/2.31 \\
Total (4.0s) & 0.71/0.87 & 1.78/2.90 & 1.71/2.98 \\
\bottomrule
\end{tabular}
}
\vspace{-0.1cm}
\end{table}


\subsection{Qualitative Results}
We illustrate our model's performance by visualizing the predicted trajectories on the NBA sports dataset and ETH-UCY pedestrian dataset. Consider ~\cref{fig:qualresult_accuracy_nba}, our \model IMLE predicts more accurate results than the LED model, which aligns to the results in ~\cref{tab:nba_result}. In~\cref{fig:qualresult_diversity}, the diverse predicted trajectories of professional basketball player and basketball are showcased. Likewise, as demonstrated in the scenarios from the ETH-UCY dataset in~\cref{fig:qualresult_accuracy_eth}, our predictions frequently align with the ground-truth pedestrain trajectories. Note that we employ two versions of the ETH-UCY dataset. The results reported in~\cref{tab:eth_ucy_result} are based on the dataset versions utilized in MID~\cite{Gu_2022_CVPR} and LED~\cite{Mao_2023_CVPR}. In contrast, we present the qualitative result~\cref{fig:qualresult_accuracy_eth} using the ETH-UCY test set from SocialGAN\cite{gupta2018social}, following the same data split. Moreover, the quantitative results for this version are provided in 
the Supplementary Material.

\subsection{Ablation Studies}

Regarding the IMLE objective, we conduct a comparative analysis using DCGAN~\cite{radford2016unsupervised} and WGAN~\cite{arjovsky2017wasserstein} as baseline implicit models in~\cref{tab:ablation_nba}.
GAN-based distillation methods fail to deliver satisfactory performance due to their inherent training instability.
In contrast, the IMLE training objective is both fast and stable, achieving state-of-the-art performance compared to existing baselines, as shown in~\cref{tab:nba_result}.

\section{Conclusion}
This paper presents a novel \model model for trajectory prediction tasks. Our new flow matching loss effectively learns a diverse set of future trajectories, addressing the inherent multi-modality of human motion.
Additionally, we propose the first one-step IMLE distillation model, which involves a process to strategically match the multi-modal future trajectories through a latent variable. This distillation method ameliorates the trilemma~\cite{xiao2022DDGAN} in generative learning, enabling the fast generation of accurate and diverse future trajectories. Our approach demonstrates strong performance, delivering results that are competitive with, and in some aspects surpass, those of current state-of-the-art models, as evidenced by our evaluations on the NBA SportVU, ETH-UCY and SDD datasets. Our distilled student model achieves a 100-fold increase in sampling speed due to its one-step prediction mechanism.
This underscores the potential of the IMLE distillation model as a step forward in the field of stochastic trajectory forecasting, providing both efficiency and effectiveness in generative learning tasks.
\newline
\textbf{Future work.} One possible direction is to leverage semantic maps and scene graphs to encode environmental features such as lanes, sidewalks, and obstacles directly into the model. These structured representations enrich contextual understanding, improving the model’s capability to generate trajectories that are both context-aware and constraint-compliant. Furthermore, this approach can be extended to large-scale autonomous driving datasets, broadening its applicability and robustness.

\clearpage

\section*{Acknowledgment}
This work was funded, in part, by the NSERC Discovery Grants (No. RGPIN-2022-04636, RGPIN-2019-05448, and RGPIN-2021-04012), the Vector Institute for AI, and Canada CIFAR AI Chair. 
Resources used in preparing this research were provided, in part, by the Province of Ontario, the Government of Canada through the Digital Research Alliance of Canada \url{alliance.can.ca}, and companies sponsoring the Vector Institute \url{www.vectorinstitute.ai/#partners}, and Advanced Research Computing at University of British Columbia. 
Additional hardware support was provided by John R. Evans Leaders Fund CFI grant.

\bibliographystyle{ieeenat_fullname}
\bibliography{main}

@String(CVPR= {IEEE Conf. Comput. Vis. Pattern Recog.})

@String(ICCV= {Int. Conf. Comput. Vis.})

@String(ICLR = {Int. Conf. Learn. Represent.})

@String(AAAI = {AAAI})

@String(CVPR  = {CVPR})

@String(ICCV  = {ICCV})

@String(ICLR  = {ICLR})

@article{vaswani2017attention,
  title={Attention is all you need},
  author={Vaswani, Ashish and Shazeer, Noam and Parmar, Niki and Uszkoreit, Jakob and Jones, Llion and Gomez, Aidan N and Kaiser, {\L}ukasz and Polosukhin, Illia},
  journal={Advances in neural information processing systems},
  volume={30},
  year={2017}
}

@InProceedings{groupnet_CVPR,
    author    = {Xu, Chenxin and Li, Maosen and Ni, Zhenyang and Zhang, Ya and Chen, Siheng},
    title     = {GroupNet: Multiscale Hypergraph Neural Networks for Trajectory Prediction With Relational Reasoning},
    booktitle = {Proceedings of the IEEE/CVF Conference on Computer Vision and Pattern Recognition (CVPR)},
    month     = {June},
    year      = {2022},
    pages     = {6498-6507}
}

@InProceedings{Bae_2022_CVPR,
    author    = {Bae, Inhwan and Park, Jin-Hwi and Jeon, Hae-Gon},
    title     = {Non-Probability Sampling Network for Stochastic Human Trajectory Prediction},
    booktitle = {Proceedings of the IEEE/CVF Conference on Computer Vision and Pattern Recognition (CVPR)},
    month     = {June},
    year      = {2022},
    pages     = {6477-6487}
}

@InProceedings{memonet_CVPR,
    author    = {Xu, Chenxin and Mao, Weibo and Zhang, Wenjun and Chen, Siheng},
    title     = {Remember Intentions: Retrospective-Memory-Based Trajectory Prediction},
    booktitle = {Proceedings of the IEEE/CVF Conference on Computer Vision and Pattern Recognition (CVPR)},
    month     = {June},
    year      = {2022},
    pages     = {6488-6497}
}

@InProceedings{Mao_2023_CVPR,
    author    = {Mao, Weibo and Xu, Chenxin and Zhu, Qi and Chen, Siheng and Wang, Yanfeng},
    title     = {Leapfrog Diffusion Model for Stochastic Trajectory Prediction},
    booktitle = {Proceedings of the IEEE/CVF Conference on Computer Vision and Pattern Recognition (CVPR)},
    month     = {June},
    year      = {2023},
    pages     = {5517-5526}
}

@article{CAGN_2022, title={Complementary Attention Gated Network for Pedestrian Trajectory Prediction}, volume={36}, url={https://ojs.aaai.org/index.php/AAAI/article/view/19933}, DOI={10.1609/aaai.v36i1.19933}, number={1}, journal={Proceedings of the AAAI Conference on Artificial Intelligence}, author={Duan, Jinghai and Wang, Le and Long, Chengjiang and Zhou, Sanping and Zheng, Fang and Shi, Liushuai and Hua, Gang}, year={2022}, month={Jun.}, pages={542-550} }

@inproceedings{mangalam2021goals,
 author = {Mangalam, Karttikeya and An, Yang and Girase, Harshayu and Malik, Jitendra},
 title = {From Goals, Waypoints \& Paths To Long Term Human Trajectory Forecasting},
 booktitle = {Proc. International Conference on Computer Vision (ICCV)},
 year = {2021},
 month = oct,
 month_numeric = {10}
}

@inproceedings{li2020Evolvegraph,
  title={EvolveGraph: Multi-Agent Trajectory Prediction with Dynamic Relational Reasoning},
  author={Li, Jiachen and Yang, Fan and Tomizuka, Masayoshi and Choi, Chiho},
  booktitle={Proceedings of the Neural Information Processing Systems (NeurIPS)},
  year={2020}
}

@inproceedings{socialvae2022,
   author={Xu, Pei and Hayet, Jean-Bernard and Karamouzas, Ioannis},
   title={SocialVAE: Human Trajectory Prediction using Timewise Latents},
   booktitle={European Conference on Computer Vision},
   pages={511-528},
   year={2022},
   organization={Springer},
   doi={10.1007/978-3-031-19772-7_30}
}

@inproceedings{xiao2022DDGAN,
    title={Tackling the Generative Learning Trilemma with Denoising Diffusion {GAN}s},
    author={Zhisheng Xiao and Karsten Kreis and Arash Vahdat},
    booktitle={International Conference on Learning Representations (ICLR)},
    year={2022}
}

@article{ho2020denoising,
  title={Denoising diffusion probabilistic models},
  author={Ho, Jonathan and Jain, Ajay and Abbeel, Pieter},
  journal={Advances in neural information processing systems},
  volume={33},
  pages={6840--6851},
  year={2020}
}

@article{song2020denoising,
  title={Denoising diffusion implicit models},
  author={Song, Jiaming and Meng, Chenlin and Ermon, Stefano},
  journal={arXiv preprint arXiv:2010.02502},
  year={2020}
}

@article{luo2023comprehensive,
  title={A comprehensive survey on knowledge distillation of diffusion models},
  author={Luo, Weijian},
  journal={arXiv preprint arXiv:2304.04262},
  year={2023}
}

@article{luhman2021knowledge,
  title={Knowledge distillation in iterative generative models for improved sampling speed},
  author={Luhman, Eric and Luhman, Troy},
  journal={arXiv preprint arXiv:2101.02388},
  year={2021}
}

@inproceedings{gu2023boot,
  title={Boot: Data-free distillation of denoising diffusion models with bootstrapping},
  author={Gu, Jiatao and Zhai, Shuangfei and Zhang, Yizhe and Liu, Lingjie and Susskind, Joshua M},
  booktitle={ICML 2023 Workshop on Structured Probabilistic Inference and Generative Modeling},
  year={2023}
}

@article{Xie2024EMDF,
  title={EM Distillation for One-step Diffusion Models},
  author={Sirui Xie and Zhisheng Xiao and Diederik P. Kingma and Tingbo Hou and Ying Nian Wu and Kevin Patrick Murphy and Tim Salimans and Ben Poole and Ruiqi Gao},
  journal={ArXiv},
  year={2024},
  volume={abs/2405.16852},
  url={https://api.semanticscholar.org/CorpusID:270062581}
}

@article{berthelot2023tract,
  title={Tract: Denoising diffusion models with transitive closure time-distillation},
  author={Berthelot, David and Autef, Arnaud and Lin, Jierui and Yap, Dian Ang and Zhai, Shuangfei and Hu, Siyuan and Zheng, Daniel and Talbott, Walter and Gu, Eric},
  journal={arXiv preprint arXiv:2303.04248},
  year={2023}
}

@InProceedings{Kim_2024_CVPR,
    author    = {Kim, Sungjune and Chi, Hyung-gun and Lim, Hyerin and Ramani, Karthik and Kim, Jinkyu and Kim, Sangpil},
    title     = {Higher-order Relational Reasoning for Pedestrian Trajectory Prediction},
    booktitle = {Proceedings of the IEEE/CVF Conference on Computer Vision and Pattern Recognition (CVPR)},
    month     = {June},
    year      = {2024},
    pages     = {15251-15260}
}

@inproceedings{
frans2025one,
title={One Step Diffusion via Shortcut Models},
author={Kevin Frans and Danijar Hafner and Sergey Levine and Pieter Abbeel},
booktitle={The Thirteenth International Conference on Learning Representations},
year={2025},
url={https://openreview.net/forum?id=OlzB6LnXcS}
}

@inproceedings{zhou2024score,
  title={Score identity distillation: Exponentially fast distillation of pretrained diffusion models for one-step generation},
  author={Zhou, Mingyuan and Zheng, Huangjie and Wang, Zhendong and Yin, Mingzhang and Huang, Hai},
  booktitle={Forty-first International Conference on Machine Learning},
  year={2024}
}

@inproceedings{yin2024one,
  title={One-step diffusion with distribution matching distillation},
  author={Yin, Tianwei and Gharbi, Micha{\"e}l and Zhang, Richard and Shechtman, Eli and Durand, Fredo and Freeman, William T and Park, Taesung},
  booktitle={Proceedings of the IEEE/CVF Conference on Computer Vision and Pattern Recognition},
  pages={6613--6623},
  year={2024}
}

@article{yin2025improved,
  title={Improved distribution matching distillation for fast image synthesis},
  author={Yin, Tianwei and Gharbi, Micha{\"e}l and Park, Taesung and Zhang, Richard and Shechtman, Eli and Durand, Fredo and Freeman, Bill},
  journal={Advances in Neural Information Processing Systems},
  volume={37},
  pages={47455--47487},
  year={2024}
}

@InProceedings{Gu_2022_CVPR,
    author    = {Gu, Tianpei and Chen, Guangyi and Li, Junlong and Lin, Chunze and Rao, Yongming and Zhou, Jie and Lu, Jiwen},
    title     = {Stochastic Trajectory Prediction via Motion Indeterminacy Diffusion},
    booktitle = {Proceedings of the IEEE/CVF Conference on Computer Vision and Pattern Recognition (CVPR)},
    month     = {June},
    year      = {2022},
    pages     = {17113-17122}
}

@inproceedings{yuan2021agent,
    title={AgentFormer: Agent-Aware Transformers for Socio-Temporal Multi-Agent Forecasting},
    author={Yuan, Ye and Weng, Xinshuo and Ou, Yanglan and Kitani, Kris},
    booktitle={Proceedings of the IEEE/CVF International Conference on Computer Vision (ICCV)},
    year={2021}
}

@inproceedings{ngiamscene,
  title={Scene Transformer: A unified architecture for predicting future trajectories of multiple agents},
  author={Ngiam, Jiquan and Vasudevan, Vijay and Caine, Benjamin and Zhang, Zhengdong and Chiang, Hao-Tien Lewis and Ling, Jeffrey and Roelofs, Rebecca and Bewley, Alex and Liu, Chenxi and Venugopal, Ashish and others},
  booktitle={International Conference on Learning Representations},
year = {2021}
}

@inproceedings{qi2017pointnet,
  title={Pointnet: Deep learning on point sets for 3d classification and segmentation},
  author={Qi, Charles R and Su, Hao and Mo, Kaichun and Guibas, Leonidas J},
  booktitle={Proceedings of the IEEE conference on computer vision and pattern recognition},
  pages={652--660},
  year={2017}
}

@article{li2018implicit,
  title={Implicit maximum likelihood estimation},
  author={Li, Ke and Malik, Jitendra},
  journal={arXiv preprint arXiv:1809.09087},
  year={2018}
}

@inproceedings{li2019diverse,
  title={Diverse Image Synthesis from Semantic Layouts via Conditional IMLE},
  author={Ke Li and Tianhao Zhang and Jitendra Malik},  
  booktitle={Proceedings of the IEEE international conference on computer vision},
  year={2019}
}

@inproceedings{sauer2024adversarial,
  title={Adversarial diffusion distillation},
  author={Sauer, Axel and Lorenz, Dominik and Blattmann, Andreas and Rombach, Robin},
  booktitle={European Conference on Computer Vision},
  pages={87--103},
  year={2024},
  organization={Springer}
}

@inproceedings{
  song2021scorebased,
  title={Score-Based Generative Modeling through Stochastic Differential Equations},
  author={Yang Song and Jascha Sohl-Dickstein and Diederik P Kingma and Abhishek Kumar and Stefano Ermon and Ben Poole},
  booktitle={International Conference on Learning Representations},
  year={2021},
  url={https://openreview.net/forum?id=PxTIG12RRHS}
}

@inproceedings{
  PD,
  title={Progressive Distillation for Fast Sampling of Diffusion Models},
  author={Tim Salimans, Jonathan Ho},
  booktitle={International Conference on Learning Representations},
  year={2021}
}

@InProceedings{Meng_2023_CVPR,
    author    = {Meng, Chenlin and Rombach, Robin and Gao, Ruiqi and Kingma, Diederik and Ermon, Stefano and Ho, Jonathan and Salimans, Tim},
    title     = {On Distillation of Guided Diffusion Models},
    booktitle = {Proceedings of the IEEE/CVF Conference on Computer Vision and Pattern Recognition (CVPR)},
    month     = {June},
    year      = {2023},
    pages     = {14297-14306}
}

@InProceedings{sohl-dickstein15,
  title = 	 {Deep Unsupervised Learning using Nonequilibrium Thermodynamics},
  author = 	 {Sohl-Dickstein, Jascha and Weiss, Eric and Maheswaranathan, Niru and Ganguli, Surya},
  booktitle = 	 {Proceedings of the 32nd International Conference on Machine Learning},
  pages = 	 {2256--2265},
  year = 	 {2015},
  editor = 	 {Bach, Francis and Blei, David},
  volume = 	 {37},
  series = 	 {Proceedings of Machine Learning Research},
  address = 	 {Lille, France},
  month = 	 {07--09 Jul},
}

@inproceedings{song2019nips,
 author = {Song, Yang and Ermon, Stefano},
 booktitle = {Advances in Neural Information Processing Systems},
 editor = {H. Wallach and H. Larochelle and A. Beygelzimer and F. d\textquotesingle Alch\'{e}-Buc and E. Fox and R. Garnett},
 pages = {},
 publisher = {Curran Associates, Inc.},
 title = {Generative Modeling by Estimating Gradients of the Data Distribution},
 url = {https://proceedings.neurips.cc/paper_files/paper/2019/file/3001ef257407d5a371a96dcd947c7d93-Paper.pdf},
 volume = {32},
 year = {2019}
}

@inproceedings{bar2024lumiere,
  title={Lumiere: A space-time diffusion model for video generation},
  author={Bar-Tal, Omer and Chefer, Hila and Tov, Omer and Herrmann, Charles and Paiss, Roni and Zada, Shiran and Ephrat, Ariel and Hur, Junhwa and Liu, Guanghui and Raj, Amit and others},
  booktitle={SIGGRAPH Asia 2024 Conference Papers},
  pages={1--11},
  year={2024}
}

@article{ho2022video,
  title={Video diffusion models},
  author={Ho, Jonathan and Salimans, Tim and Gritsenko, Alexey and Chan, William and Norouzi, Mohammad and Fleet, David J},
  journal={Advances in Neural Information Processing Systems},
  volume={35},
  pages={8633--8646},
  year={2022}
}

@inproceedings{zeng2022lion,
    title={LION: Latent Point Diffusion Models for 3D Shape Generation},
        author={Xiaohui Zeng and Arash Vahdat and Francis Williams and Zan Gojcic and Or Litany and Sanja Fidler and Karsten Kreis},
        booktitle={Advances in Neural Information Processing Systems (NeurIPS)},
        year={2022}
}

@inproceedings{ettinger2021large,
  title={Large scale interactive motion forecasting for autonomous driving: The waymo open motion dataset},
  author={Ettinger, Scott and Cheng, Shuyang and Caine, Benjamin and Liu, Chenxi and Zhao, Hang and Pradhan, Sabeek and Chai, Yuning and Sapp, Ben and Qi, Charles R and Zhou, Yin and others},
  booktitle={Proceedings of the IEEE/CVF International Conference on Computer Vision},
  pages={9710--9719},
  year={2021}
}

@article{floreano2015science,
  title={Science, technology and the future of small autonomous drones},
  author={Floreano, Dario and Wood, Robert J},
  journal={nature},
  volume={521},
  number={7553},
  pages={460--466},
  year={2015},
  publisher={Nature Publishing Group UK London}
}

@article{zhang2020recurrent,
  title={Recurrent neural network for motion trajectory prediction in human-robot collaborative assembly},
  author={Zhang, Jianjing and Liu, Hongyi and Chang, Qing and Wang, Lihui and Gao, Robert X},
  journal={CIRP annals},
  volume={69},
  number={1},
  pages={9--12},
  year={2020},
  publisher={Elsevier}
}

@inproceedings{nayakanti2023wayformer,
  title={Wayformer: Motion forecasting via simple \& efficient attention networks},
  author={Nayakanti, Nigamaa and Al-Rfou, Rami and Zhou, Aurick and Goel, Kratarth and Refaat, Khaled S and Sapp, Benjamin},
  booktitle={2023 IEEE International Conference on Robotics and Automation (ICRA)},
  pages={2980--2987},
  year={2023},
  organization={IEEE}
}

@InProceedings{Wu_2023_CVPR,
    author    = {Wu, Lemeng and Wang, Dilin and Gong, Chengyue and Liu, Xingchao and Xiong, Yunyang and Ranjan, Rakesh and Krishnamoorthi, Raghuraman and Chandra, Vikas and Liu, Qiang},
    title     = {Fast Point Cloud Generation With Straight Flows},
    booktitle = {Proceedings of the IEEE/CVF Conference on Computer Vision and Pattern Recognition (CVPR)},
    month     = {June},
    year      = {2023},
    pages     = {9445-9454}
}

@article{bhattacharyya2019conditional,
  title={Conditional flow variational autoencoders for structured sequence prediction},
  author={Bhattacharyya, Apratim and Hanselmann, Michael and Fritz, Mario and Schiele, Bernt and Straehle, Christoph-Nikolas},
  journal={arXiv preprint arXiv:1908.09008, Bayesian Deep Learning and Machine Learning for Autonomous Driving NeurIPS},
  year={2019}
}

@misc{lu2023dpmsolverfastsolverguided,
      title={DPM-Solver++: Fast Solver for Guided Sampling of Diffusion Probabilistic Models}, 
      author={Cheng Lu and Yuhao Zhou and Fan Bao and Jianfei Chen and Chongxuan Li and Jun Zhu},
      year={2023},
      eprint={2211.01095},
      archivePrefix={arXiv},
      primaryClass={cs.LG},
      url={https://arxiv.org/abs/2211.01095}, 
}

@inproceedings{liu2023instaflow,
  title={Instaflow: One step is enough for high-quality diffusion-based text-to-image generation},
  author={Liu, Xingchao and Zhang, Xiwen and Ma, Jianzhu and Peng, Jian and Liu, Qiang},
  booktitle={International Conference on Learning Representations},
  year={2024}
}

@inproceedings{zheng2023fastsampling,
  archiveprefix = {arxiv},
  eprint        = {2211.13449},
  title         = {Fast Sampling of Diffusion Models via Operator Learning},
  author        = {Hongkai Zheng and Weili Nie and Arash Vahdat and Kamyar Azizzadenesheli and Anima Anandkumar},
  booktitle     = {International Conference on Machine Learning (ICML)},
  month         = {July},
  year          = {2023},
}

@article{kingma2018glow,
  title={Glow: Generative flow with invertible 1x1 convolutions},
  author={Kingma, Durk P and Dhariwal, Prafulla},
  journal={Advances in neural information processing systems},
  volume={31},
  year={2018}
}

@inproceedings{
Yuan2020Diverse,
title={Diverse Trajectory Forecasting with Determinantal Point Processes},
author={Ye Yuan and Kris M. Kitani},
booktitle={International Conference on Learning Representations},
year={2020},
url={https://openreview.net/forum?id=ryxnY3NYPS}
}

@inproceedings{
kim2025simple,
title={Simple ReFlow: Improved Techniques for Fast Flow Models},
author={Beomsu Kim and Yu-Guan Hsieh and Michal Klein and marco cuturi and Jong Chul Ye and Bahjat Kawar and James Thornton},
booktitle={The Thirteenth International Conference on Learning Representations},
year={2025},
url={https://openreview.net/forum?id=fpvgSDKXGY}
}

@article{luo2023diff,
  title={Diff-instruct: A universal approach for transferring knowledge from pre-trained diffusion models},
  author={Luo, Weijian and Hu, Tianyang and Zhang, Shifeng and Sun, Jiacheng and Li, Zhenguo and Zhang, Zhihua},
  journal={Advances in Neural Information Processing Systems},
  volume={36},
  pages={76525--76546},
  year={2023}
}

@article{wang2023prolificdreamer,
  title={Prolificdreamer: High-fidelity and diverse text-to-3d generation with variational score distillation},
  author={Wang, Zhengyi and Lu, Cheng and Wang, Yikai and Bao, Fan and Li, Chongxuan and Su, Hang and Zhu, Jun},
  journal={Advances in Neural Information Processing Systems},
  volume={36},
  pages={8406--8441},
  year={2023}
}

@inproceedings{weng2023joint,
  title={Joint metrics matter: A better standard for trajectory forecasting},
  author={Weng, Erica and Hoshino, Hana and Ramanan, Deva and Kitani, Kris},
  booktitle={Proceedings of the IEEE/CVF International Conference on Computer Vision},
  pages={20315--20326},
  year={2023}
}

@inproceedings{wong2024socialcircle,
  title={SocialCircle: Learning the Angle-based Social Interaction Representation for Pedestrian Trajectory Prediction},
  author={Wong, Conghao and Xia, Beihao and Zou, Ziqian and Wang, Yulong and You, Xinge},
  booktitle={Proceedings of the IEEE/CVF Conference on Computer Vision and Pattern Recognition},
  pages={19005--19015},
  year={2024}
}

@inproceedings{lin2024infoCD,
 author = {Lin, Fangzhou and Yue, Yun and Zhang, Ziming and Hou, Songlin and Yamada, Kazunori and Kolachalama, Vijaya and Saligrama, Venkatesh},
 booktitle = {Advances in Neural Information Processing Systems},
 editor = {A. Oh and T. Naumann and A. Globerson and K. Saenko and M. Hardt and S. Levine},
 pages = {76960--76973},
 publisher = {Curran Associates, Inc.},
 title = {InfoCD: A Contrastive Chamfer Distance Loss for Point Cloud Completion},
 url = {https://proceedings.neurips.cc/paper_files/paper/2023/file/f2ea1943896474b7cd9796b93e526f6f-Paper-Conference.pdf},
 volume = {36},
 year = {2023}
}

@inproceedings{
kollovieh2023predict,
title={Predict, Refine, Synthesize: Self-Guiding Diffusion Models for Probabilistic Time Series Forecasting},
author={Marcel Kollovieh and Abdul Fatir Ansari and Michael Bohlke-Schneider and Jasper Zschiegner and Hao Wang and Bernie Wang},
booktitle={Thirty-seventh Conference on Neural Information Processing Systems},
year={2023},
url={https://openreview.net/forum?id=q6X038vKgU}
}

@article{hinton2015distillingknowledgeneuralnetwork,
  title={Distilling the knowledge in a neural network},
  author={Hinton, Geoffrey and Vinyals, Oriol and Dean, Jeff},
  journal={arXiv preprint arXiv:1503.02531},
  year={2015}
}

@article{radford2016unsupervised,
  title={Unsupervised representation learning with deep convolutional generative adversarial networks},
  author={Radford, Alec and Metz, Luke and Chintala, Soumith},
  journal={arXiv preprint arXiv:1511.06434},
  year={2016}
}

@inproceedings{arjovsky2017wasserstein,
  title={Wasserstein generative adversarial networks},
  author={Arjovsky, Martin and Chintala, Soumith and Bottou, L{\'e}on},
  booktitle={International conference on machine learning},
  pages={214--223},
  year={2017},
  organization={PMLR}
}

@InProceedings{Shi_2023_ICCV,
    author    = {Shi, Liushuai and Wang, Le and Zhou, Sanping and Hua, Gang},
    title     = {Trajectory Unified Transformer for Pedestrian Trajectory Prediction},
    booktitle = {Proceedings of the IEEE/CVF International Conference on Computer Vision (ICCV)},
    month     = {October},
    year      = {2023},
    pages     = {9675-9684}
}

@inproceedings{bae2024singulartrajectory,
  title={SingularTrajectory: Universal Trajectory Predictor Using Diffusion Model},
  author={Bae, Inhwan and Park, Young-Jae and Jeon, Hae-Gon},
  booktitle={Proceedings of the IEEE/CVF Conference on Computer Vision and Pattern Recognition},
  year={2024}
}

@inproceedings{xu2023eqmotion,
  title={EqMotion: Equivariant Multi-agent Motion Prediction with Invariant Interaction Reasoning},
  author={Xu, Chenxin and Tan, Robby T and Tan, Yuhong and Chen, Siheng and Wang, Yu Guang and Wang, Xinchao and Wang, Yanfeng},
  booktitle={Proceedings of the IEEE/CVF Conference on Computer Vision and Pattern Recognition},
  pages={1410--1420},
  year={2023}
}

@inproceedings{bae2023eigentrajectory,
  title={EigenTrajectory: Low-Rank Descriptors for Multi-Modal Trajectory Forecasting},
  author={Bae, Inhwan and Oh, Jean and Jeon, Hae-Gon},
  booktitle={Proceedings of the IEEE/CVF International Conference on Computer Vision},
  year={2023}
}

@inproceedings{dendorfer2021mg,
  title={Mg-gan: A multi-generator model preventing out-of-distribution samples in pedestrian trajectory prediction},
  author={Dendorfer, Patrick and Elflein, Sven and Leal-Taix{\'e}, Laura},
  booktitle={Proceedings of the IEEE/CVF International Conference on Computer Vision},
  pages={13158--13167},
  year={2021}
}

@article{xu2024dynamic,
  title={Dynamic-group-aware networks for multi-agent trajectory prediction with relational reasoning},
  author={Xu, Chenxin and Wei, Yuxi and Tang, Bohan and Yin, Sheng and Zhang, Ya and Chen, Siheng and Wang, Yanfeng},
  journal={Neural Networks},
  volume={170},
  pages={564--577},
  year={2024},
  publisher={Elsevier}
}

@InProceedings{Lee_2022_CVPR,
    author    = {Lee, Mihee and Sohn, Samuel S. and Moon, Seonghyeon and Yoon, Sejong and Kapadia, Mubbasir and Pavlovic, Vladimir},
    title     = {MUSE-VAE: Multi-Scale VAE for Environment-Aware Long Term Trajectory Prediction},
    booktitle = {Proceedings of the IEEE/CVF Conference on Computer Vision and Pattern Recognition (CVPR)},
    month     = {June},
    year      = {2022},
    pages     = {2221-2230}
}

@misc{wu2024consistent3d,
      title={Consistent3D: Towards Consistent High-Fidelity Text-to-3D Generation with Deterministic Sampling Prior}, 
      author={Zike Wu and Pan Zhou and Xuanyu Yi and Xiaoding Yuan and Hanwang Zhang},
      year={2024},
      eprint={2401.09050},
      archivePrefix={arXiv},
      primaryClass={cs.CV}
}

@article{wu2023fast,
  title={Fast diffusion model},
  author={Wu, Zike and Zhou, Pan and Kawaguchi, Kenji and Zhang, Hanwang},
  journal={arXiv preprint arXiv:2306.06991},
  year={2023}
}

@article{liu2022flow,
  title={Flow straight and fast: Learning to generate and transfer data with rectified flow},
  author={Liu, Xingchao and Gong, Chengyue and Liu, Qiang},
  journal={arXiv preprint arXiv:2209.03003},
  year={2022}
}

@article{lipman2022flow,
  title={Flow matching for generative modeling},
  author={Lipman, Yaron and Chen, Ricky TQ and Ben-Hamu, Heli and Nickel, Maximilian and Le, Matt},
  journal={arXiv preprint arXiv:2210.02747},
  year={2022}
}

@inproceedings{esser2024scaling,
  title={Scaling rectified flow transformers for high-resolution image synthesis},
  author={Esser, Patrick and Kulal, Sumith and Blattmann, Andreas and Entezari, Rahim and M{\"u}ller, Jonas and Saini, Harry and Levi, Yam and Lorenz, Dominik and Sauer, Axel and Boesel, Frederic and others},
  booktitle={Forty-first International Conference on Machine Learning},
  year={2024}
}

@article{albergo2023stochastic,
  title={Stochastic interpolants: A unifying framework for flows and diffusions},
  author={Albergo, Michael S and Boffi, Nicholas M and Vanden-Eijnden, Eric},
  journal={arXiv preprint arXiv:2303.08797},
  year={2023}
}

@inproceedings{jiang2023motiondiffuser,
  title={Motiondiffuser: Controllable multi-agent motion prediction using diffusion},
  author={Jiang, Chiyu and Cornman, Andre and Park, Cheolho and Sapp, Benjamin and Zhou, Yin and Anguelov, Dragomir and others},
  booktitle={Proceedings of the IEEE/CVF Conference on Computer Vision and Pattern Recognition},
  pages={9644--9653},
  year={2023}
}

@article{gat2025discrete,
  title={Discrete flow matching},
  author={Gat, Itai and Remez, Tal and Shaul, Neta and Kreuk, Felix and Chen, Ricky TQ and Synnaeve, Gabriel and Adi, Yossi and Lipman, Yaron},
  journal={Advances in Neural Information Processing Systems},
  volume={37},
  pages={133345--133385},
  year={2024}
}

@article{polyak2024movie,
  title={Movie gen: A cast of media foundation models},
  author={Polyak, Adam and Zohar, Amit and Brown, Andrew and Tjandra, Andros and Sinha, Animesh and Lee, Ann and Vyas, Apoorv and Shi, Bowen and Ma, Chih-Yao and Chuang, Ching-Yao and others},
  journal={arXiv preprint arXiv:2410.13720},
  year={2024}
}

@inproceedings{liu2021social,
  title={Social nce: Contrastive learning of socially-aware motion representations},
  author={Liu, Yuejiang and Yan, Qi and Alahi, Alexandre},
  booktitle={Proceedings of the IEEE/CVF International Conference on Computer Vision},
  pages={15118--15129},
  year={2021}
}

@inproceedings{luo2021safety,
  title={Safety-oriented pedestrian occupancy forecasting},
  author={Luo, Katie and Casas, Sergio and Liao, Renjie and Yan, Xinchen and Xiong, Yuwen and Zeng, Wenyuan and Urtasun, Raquel},
  booktitle={2021 IEEE/RSJ International Conference on Intelligent Robots and Systems (IROS)},
  pages={1015--1022},
  year={2021},
  organization={IEEE}
}

@inproceedings{jain2020discrete,
  title={Discrete residual flow for probabilistic pedestrian behavior prediction},
  author={Jain, Ajay and Casas, Sergio and Liao, Renjie and Xiong, Yuwen and Feng, Song and Segal, Sean and Urtasun, Raquel},
  booktitle={Conference on Robot Learning},
  pages={407--419},
  year={2020},
  organization={PMLR}
}

@article{song2023consistency,
  title={Consistency models},
  author={Song, Yang and Dhariwal, Prafulla and Chen, Mark and Sutskever, Ilya},
  journal={arXiv preprint arXiv:2303.01469},
  year={2023}
}

@inproceedings{gupta2018social,
  title={Social gan: Socially acceptable trajectories with generative adversarial networks},
  author={Gupta, Agrim and Johnson, Justin and Fei-Fei, Li and Savarese, Silvio and Alahi, Alexandre},
  booktitle={Proceedings of the IEEE conference on computer vision and pattern recognition},
  pages={2255--2264},
  year={2018}
}

@inproceedings{sadeghian2019sophie,
  title={Sophie: An attentive gan for predicting paths compliant to social and physical constraints},
  author={Sadeghian, Amir and Kosaraju, Vineet and Sadeghian, Ali and Hirose, Noriaki and Rezatofighi, Hamid and Savarese, Silvio},
  booktitle={Proceedings of the IEEE/CVF conference on computer vision and pattern recognition},
  pages={1349--1358},
  year={2019}
}

@article{melnik2024videodiffusion,
  title={Video Diffusion Models: A Survey},
  author={Melnik, Andrew and Ljubljanac, Michal and Lu, Cong and Yan, Qi and Ren, Weiming and Ritter, Helge},
  journal={arXiv preprint arXiv:2405.03150},
  year={2024}
}

@article{yan2023swingnn,
  title={Swingnn: Rethinking permutation invariance in diffusion models for graph generation},
  author={Yan, Qi and Liang, Zhengyang and Song, Yang and Liao, Renjie and Wang, Lele},
  journal={arXiv preprint arXiv:2307.01646},
  year={2023}
}

@article{abramson2024accurate,
  title={Accurate structure prediction of biomolecular interactions with AlphaFold 3},
  author={Abramson, Josh and Adler, Jonas and Dunger, Jack and Evans, Richard and Green, Tim and Pritzel, Alexander and Ronneberger, Olaf and Willmore, Lindsay and Ballard, Andrew J and Bambrick, Joshua and others},
  journal={Nature},
  pages={1--3},
  year={2024},
  publisher={Nature Publishing Group UK London}
}

@inproceedings{suo2021trafficsim,
  title={Trafficsim: Learning to simulate realistic multi-agent behaviors},
  author={Suo, Simon and Regalado, Sebastian and Casas, Sergio and Urtasun, Raquel},
  booktitle={Proceedings of the IEEE/CVF Conference on Computer Vision and Pattern Recognition},
  pages={10400--10409},
  year={2021}
}

@inproceedings{
loshchilov2018decoupled,
title={Decoupled Weight Decay Regularization},
author={Ilya Loshchilov and Frank Hutter},
booktitle={International Conference on Learning Representations},
year={2019},
url={https://openreview.net/forum?id=Bkg6RiCqY7},
}

@article{paszke2019pytorch,
  title={Pytorch: An imperative style, high-performance deep learning library},
  author={Paszke, Adam and Gross, Sam and Massa, Francisco and Lerer, Adam and Bradbury, James and Chanan, Gregory and Killeen, Trevor and Lin, Zeming and Gimelshein, Natalia and Antiga, Luca and others},
  journal={Advances in neural information processing systems},
  volume={32},
  year={2019}
}

\clearpage
\maketitlesupplementary
\appendix

\begin{figure*}[htbp] 
    \vspace{-0.5cm}
    \begin{minipage}[t]{0.48\textwidth} 
    \begin{algorithm}[H] 
    \caption{\model Teacher Model Training}   
    \label{alg:teacher_train}
    \begin{algorithmic}[1]
        \State\textbf{Require:}\xspace Training dataset $p_\mathcal{D}$, learning rate $\eta$, teacher model $D_\theta$ for data prediction
        \State Initialize the parameters $\theta$ of the network $D_\theta$.
        \Repeat 
            \State $(Y^1, C) \sim p_\mathcal{D}$ \Comment{Sample observed data}
            \State $Y^0_s \sim \mathcal{N}(\b 0, \b I)$
            \vspace{0.05cm}
            \State $Y^0_{1:K} \leftarrow \texttt{repeat}(Y^0_s, K) $ \Comment{Tied noise}
            \vspace{0.05cm}
            \State $Y^1_{1:K} \leftarrow \texttt{repeat}(Y^1, K) $
            \vspace{0.05cm}
            \State $t \sim p_t(t)$ \Comment{Time scheduler}
            \State $Y^t_{1:K} \leftarrow t Y^1_{1:K} + (1-t) Y^0_{1:K}$
            \vspace{0.05cm}
            \State $S_{1:K}, \zeta_{1:K} \leftarrow D_\theta(Y^t_{1:K}, C, t)$    
            \State $\theta\gets\theta-\eta\nabla_{\theta} \bar{\mathcal{L}}_{\text{FM}}(S_{1:K}, \zeta_{1:K}, Y^1)$ \Comment{\cref{eq:loss_fm_reg_cls_app}}
        \Until{converged}
        \end{algorithmic}
    \end{algorithm}
    \end{minipage}\hfill
    \begin{minipage}[t]{0.48\textwidth} 
        \begin{algorithm}[H] 
            \caption{\model Teacher Model Sampling}   
            \label{alg:teacher_sample}
            \begin{algorithmic}[1]
                \State \textbf{Require:} Evaluation dataset \(p_\mathcal{D}'\), teacher model \(D_\theta\) for data prediction, sampling steps \(T\)
                \Repeat 
                    \State \((\cdot, C) \sim p_\mathcal{D}'\) \Comment{Sample context data}
                    \State \(Y^0_s \sim \mathcal{N}(\mathbf{0}, \mathbf{I})\)
                    \vspace{0.05cm}
                    \State \(Y^0_{1:K} \gets \texttt{repeat}(Y^0_s, K)\) \Comment{Tied noise}
                    \For {\(n \in [0, 1, \cdots, T-1]\)}
                        \State \(S_{1:K}, \zeta_{1:K} \gets D_\theta(Y^{\tau_n}_{1:K}, C, \tau_n)\)
                        \vspace{0.05cm}
                        \State \(v_\theta^{(i)} = \frac{S_i - Y^{\tau_n}_i}{1 - \tau_n}, \quad \forall i \in [K]\)
                        \vspace{0.05cm}
                        \State \(v_\theta = \texttt{concat}\left(v_\theta^{(1)}, v_\theta^{(2)}, \dots, v_\theta^{(K)}\right)\)
                        \State \(Y^{\tau_{n+1}}_{1:K} \gets Y^{\tau_n}_{1:K} + (\tau_{n+1} - \tau_n) v_\theta\)
                    \EndFor
                \Until{finished}
            \end{algorithmic}
        \end{algorithm}
    \end{minipage}
    \vspace{-0.25cm}
\end{figure*}

The supplementary material is organized as follows. First, we introduce additional implementation details in~\cref{app:impl_details}, including our \model flow-matching objective in~\cref{app:fm_obj_formulation}, training and sampling steps in~\cref{app:teacher_model_training} and~\cref{app:teacher_model_sampling}, as well as the architectural details in~\cref{app:network_architecture}. In~\cref{app:qual}, we include additional qualitative results on the NBA sports dataset, demonstrating the effectiveness of our proposed methods. Moreover, we present additional ablation studies in~\cref{app:additional_exp} to showcase the success of inference-time speed-up, thanks to our IMLE distillation scheme in~\cref{app:speed_up}, and the impact of training hyperparameters of IMLE in~\cref{app:imle_ablations}.

\begin{figure*}[htbp!]
  \centering
    \includegraphics[width=\linewidth]{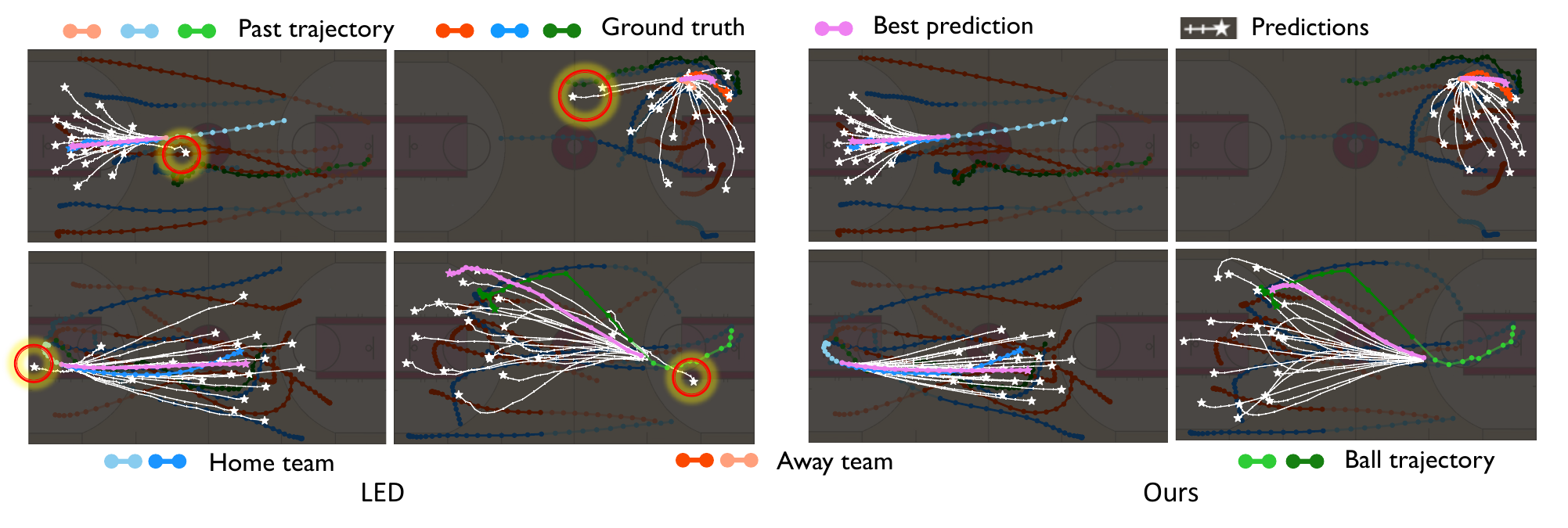}
    \vspace{-0.75cm}
    \caption{Qualitative results on NBA dataset in terms of diversity. Our method generates diverse samples that are more socially plausible. Some of the trajectories generated by LED model, which are highlighted by red circles, do not adhere to the basketball game patterns or rules. (Light color indicates past trajectory while dark color means future trajectory; blue/orange/green color: two teams and the basketball; pink color: the sample that is the closest to the Ground truth in $L_2$ sense among $K=20$ predictions)}
    \vspace{-0.5cm}
    \label{fig:supp_led_nba_20_compare}
\end{figure*}

\section{Further Implementation Details}
\label{app:impl_details}
\subsection{Flow Matching Objective Formulation}
\label{app:fm_obj_formulation}
In 
Section 4.2, we define the multi-modal motion prediction objective for the flow matching model and explain the key implementation details. 
To maintain self-consistency, we reintroduce the fundamental concepts from scratch.

Recall that our goal is to generate $K$ trajectories $\{Y_1, Y_2, \ldots, Y_K\}$ to capture the diverse motion patterns of the agents under consideration, given their overall context information denoted as $C$. 
In the flow matching framework, we extend the notation by introducing a superscript to indicate flow time, with the denoising ODE intermediate states represented as $\{Y_1^t, Y_2^t, \ldots, Y_K^t\}, t \in [0, 1]$.

However, in the dataset, we observe only a single future trajectory $Y^1$ conditioned on the context $C$. 
This limitation prevents us from directly adopting the vanilla flow matching objective, which involves linearly mixing clean data and noisy vectors to construct training objectives. 
Instead, we utilize data-space prediction and develop the multi-modal learning loss accordingly.

To achieve this, we first transform the original loss, which involves learning a vector field $v_\theta$, into predicting the data at time $t = 1$:
\begin{align*}
    \mathcal{L}_{\text{FM}} &= \mathbb{E}_{Y^t, Y^1, t} \left[ \left\Vert {v}_\theta(Y^t, C, t) - (Y^1 - Y^0) \right \Vert^2_2 \right], \\
    &= \mathbb{E}_{Y^t, Y^1, t} \left[ \left\Vert {v}_\theta(Y^t, C, t) - \frac{Y^1 - Y^t}{1-t} \right \Vert^2_2 \right], \\
    &= \mathbb{E}_{Y^t, Y^1, t} \left[ \left \Vert \frac{Y^t + (1-t){v}_\theta(Y^t, C, t) - Y^1}{1-t} \right \Vert^2_2 \right].
\end{align*}
We then define the network as a reparameterized model $D_\theta$, which implicitly learns the vector field $v_\theta$ through a linear transformation:
\begin{align*}
    &D_\theta(Y^t, C, t) \coloneqq Y^t + (1-t){v}_\theta(Y^t, C, t), \\
    &\mathcal{L}_{\text{FM}}=\mathbb{E}_{Y^t, Y^1, t} \left[ \frac{ \Vert {D}_\theta(Y^t, C, t) - Y^1 \Vert^2_2}{(1-t)^2} \right].
\end{align*}
Note that we are merely rearranging the network modules, while the loss functions remain exactly equivalent to those in the vanilla framework. 

Next, we design a training loss to encourage the data prediction model \( D_\theta \) to learn multi-modal trajectories. 
To achieve this, we employ a standard Transformer structure (as illustrated in
the first figure in the main text) to generate \( K \) correlated predictions. 
Specifically, the model \( D_\theta \) produces \( K \) scene-level waypoint predictions, denoted as 
\(\{S_i\}_{i=1}^K, \, S_i \in \mathbb{R}^{A \times 2 T_f}\), along with the corresponding classification logits \(\{\zeta_i\}_{i=1}^K, \, \zeta_i \in \mathbb{R}\).
For simplicity, we omit the time-dependent coefficients and apply a combined regression and classification loss as follows:
\begin{align}
    \bar{\mathcal{L}}_{\text{FM}} &= \mathbb{E}_{Y^t, Y^1, t} \left[\Vert S_{j^*} - Y^1 \Vert_2^2 + \mathrm{CE}(\zeta_{1:K}, j^*)\right], 
    \label{eq:loss_fm_reg_cls_app}
    \\
    j^* &= \arg \min_j \Vert S_j - Y^1 \Vert_2^2,
\end{align}
where $\mathrm{CE}(\cdot, \cdot)$ means the cross-entropy loss. 

\subsection{\model Teacher Model Training}
\label{app:teacher_model_training}
We present the training algorithm for the teacher model with a modified objective in~\cref{alg:teacher_train}. 
Notably, we observe that using tied noise across all $K$ components stabilizes the training process. In contrast, untied noise introduces excessive variability, making convergence significantly more challenging.
For the flow time scheduler $p_t(\cdot)$, we employ a logit-normal distribution: \(\text{logit}(t) \sim \mathcal{N}(\mu_t, \sigma_t^2)\). In practice, one can sample a random variable \(\kappa \sim \mathcal{N}(\mu_t, \sigma_t^2)\) and apply the standard logistic function, \(\frac{1}{1 + e^{-\kappa}}\), to obtain the desired samples for \(t\). 
The parameters \(\mu_t = -0.5\) and \(\sigma_t = 1.5\) are selected based on a hyperparameter sweep.

Empirically, we observe that the data space prediction loss function defined in~\cref{eq:loss_fm_reg_cls_app} suffers from overfitting when $t$ is close to 1.  
This is because, in this regime, the noisy input $Y^t_{1:K}$ becomes highly similar to the clean trajectory $Y^1_{1:K}$.  
When conditioned on the future trajectory $Y^1_{1:K}$, the noisy vector $Y^t_{1:K}$ at time $t$ follows a Gaussian distribution $\mathcal{N}(t Y^1_{1:K}, (1-t)^2I)$.  
Its variance, $(1-t)^2$, decreases quadratically as $t$ approaches 1.  
These overfitting effects cause the flow matching model to adopt a cheating solution during training, relying excessively on the input $Y^t_{1:K}$ and generalizing poorly on the test dataset.  
To address this issue, we propose a flow time-dependent masking mechanism to encourage the model to extract useful signals from the context $C$.  
Specifically, we introduce a masking mechanism applied to the noise embedding, as illustrated in Fig. 1
, which uses an S-shaped logistic function as the threshold:  
\[
f_m(t) = \frac{1}{1 + e^{-k (t-m)}}.
\]  
During training, for a randomly sampled time step $t$ from the time scheduler, we mask the noisy embedding with zeros with a probability of $f_m(t)$.  
This serves as a simple embedding-level dropout mechanism, applied only during training.  
In our experiments, we set $k=20$ and $m=0.5$.

\subsection{\model Teacher Model Sampling}
\label{app:teacher_model_sampling}
In~\cref{alg:teacher_sample}, we solve the denoising ODEs during sampling by leveraging the \(K\)-correlated trajectory predictions \(\{S_i\}_{i=1}^K\) produced by \(D_\theta\) to compute the \(K\)-shot vector field:
\[
v_\theta^{(i)} = \frac{S_i - Y^t_i}{1 - t}, \quad \forall i \in [K].
\]
The initial values \(Y^0_i\) for all components are sampled from a standard normal distribution. The process iteratively updates \(Y^{t}_{1:K}\) based on the predicted vector field as the flow time \(t\) progresses towards \(t=1\).

We use a mapping function \(\tau_n\) to represent the actual flow time \(t\) at sampling iteration \(n \in [0, T]\), given a total computation budget of \(T\) steps. The mapping function must satisfy two boundary conditions: \(\tau_0 = 0\) and \(\tau_{T} = 1\). In our experiments, we set \(T = 100\). Following~\cite{polyak2024movie, esser2024scaling}, we design a non-linear time mapping function as follows:
\[
\tau_n = 
\begin{cases} 
    \frac{n}{1000}, & \text{if } n \leq \frac{T}{2}, \\[8pt]
    \frac{T}{500} + \frac{\left(1.0 - \frac{T}{500}\right) \left(n - \frac{T}{2}\right)^p}{\left(\frac{T}{2}\right)^p}, & \text{if } n > \frac{T}{2}.
\end{cases}
\]
Empirically, we set \(p = 5\) based on a hyperparameter grid search.

\begin{table*}[h]
\centering
\caption{Ablation study on various components of our \model on the \textbf{NBA} dataset. (PE-K) Positional Encoding (PE) on number of predictions ($K$); (PE-A) PE on agent-level only;  (w/o PE) No PE; (Uniform) uniform noise/time schedule; (IID) i.i.d. noise instead of shared noise; (w/o Mask) Turn off the flow-time dependent masking mechanism; (Ours) Our \model with all modules on. We report min\textsubscript{20}ADE/min\textsubscript{20}FDE (meters) for empirical performance. We \textbf{bold} the top results.}
\label{tab:ablation_moflow}
\begin{tabular}{c|cccccc|c}
\toprule
\textit{Time} & PE-K & PE-A & w/o PE & Uniform & IID & w/o Mask & \textbf{Ours}\\
\midrule
1.0s & \textbf{0.19}/{0.26} & {0.41}/{0.74} & {0.45}/{0.84} & \textbf{0.19}/{0.26} & {0.26}/{0.44}& \textbf{0.19}/{0.27}&\textbf{0.19/0.25} \\
2.0s & {0.35}/\textbf{0.47} & {0.88}/{1.77} &{1.02}/{2.10} & {0.35}/{0.48} & {0.59}/{1.12}& {0.38}/{0.56}&\textbf{0.34/0.47}\\
3.0s & \textbf{0.52}/{0.68} & {1.37}/{2.72} & {1.61}/{3.22} & {0.53}/{0.68} & {0.97}/{1.87}& {0.59}/{0.80}&\textbf{0.52/0.67}\\
4.0s & \textbf{0.71}/0.88& 1.83/3.51 & 2.14/4.05 & \textbf{0.71}/0.88 & 1.36/2.46&{0.82}/{1.07} & \textbf{0.71/0.87} \\
\bottomrule
\end{tabular}
\end{table*}

\subsection{Network Architecture}
\label{app:network_architecture}
Building on prior studies~\cite{yuan2021agent, Gu_2022_CVPR, Mao_2023_CVPR}, we adopt the spatio-temporal transformer encoder to jointly model temporal and social dimensions, capturing complex agent interactions over time and space. To enhance performance across datasets, the encoder architecture is dataset-specific: the spatio-temporal transformer encoder is used for ETH-UCY and SDD, while a PointNet-like encoder~\cite{qi2017pointnet} is employed for the NBA dataset. 
Our spatio-temporal encoder processes both the context embedding and noise embedding through MLP layers, which are further combined with the flow-matching time signal during the teacher model's training. The encoder for the student IMLE model employs a similar structure, differing only in its handling of the flow-matching time. In both transformer-based encoders leverage skip connections and share a configuration of 128 features, a feed-forward dimension of 512, eight attention heads, and consist of four layers.

In our the motion decoder, we employ additional self-attention to capture interactions among $K$ scene predictions, as depicted in 
Fig 1., complementing the temporal and social interactions modeled by the encoder. The embedding dimension, feed-forward dimension, and number of attention heads in the decoder match those of the spatial temporal encoder. The decoder comprises four blocks with each block performing factorized self-attention both over the sample dimension and the agent dimension~\cite{ngiamscene}. To circumvent the potential overfitting issue, the attention dropout rate is set to 0.1.

Both our \model model and the IMLE model are smaller in size compared to the LED initializer~\cite{Mao_2023_CVPR}, with $\sim$1M fewer parameters.

\begin{table*}[t]
\vspace{-10pt}
\centering
{
\begin{tabular}{c|c|c|c|c||c|c}
\toprule
&\multicolumn{4}{c||}{$\omega=0$}& $\omega = 1$ & $\omega = 5$  \\
\hline
 & DDPM &LED & MoFlow & IMLE & MoFlow*  & MoFlow*\\
\hline
(a) @2s & 0.44/0.64 &0.37/0.57 & \textbf{ 0.34/0.47} &\textbf{0.34/0.47}  &  0.37/0.52 & 0.42/0.61\\
(a) @4s & 0.94/1.21 &0.81/1.10 & \textbf{ 0.71}/0.87 &\textbf{0.71/0.86}  &  0.75/0.96 & 0.81/1.12\\
(b) @2s & 1.15/2.30 & 0.89/1.84 &\textbf{0.83/1.64} &  \textbf{0.83}/1.65   &  0.84/1.68  &  0.86/1.73\\ 
(b) @4s & 2.40/4.65 &1.95/3.84  &    \textbf{1.71/3.34} &  1.73/3.35   &  1.74/3.37  &  1.81/3.49\\ 
\hline
 MASD~\cite{Yuan2020Diverse} & 6.41& 15.74 & 5.85 & 5.78 & 5.54 & 5.00 \\
\bottomrule
\end{tabular}
}
\caption{Joint metrics performance on NBA dataset. * means we train our MoFlow using the new objective. (a) $\min_{20}$ADE/$\min_{20}$FDE metric; (b) a new 
 metric $\min_{20}$JADE/$\min_{20}$JFDE; (c) MASD diversity metric~\cite{Yuan2020Diverse, suo2021trafficsim}. DDPM refers to the stage-one diffusion model used in LED~\cite{Mao_2023_CVPR}. We use the checkpoint from the codebase provided by LED.}
 \vspace{-10pt}
\label{tab:joint_metric}
\end{table*}

\begin{figure*}[t]
    \centering
    \includegraphics[width=\linewidth]{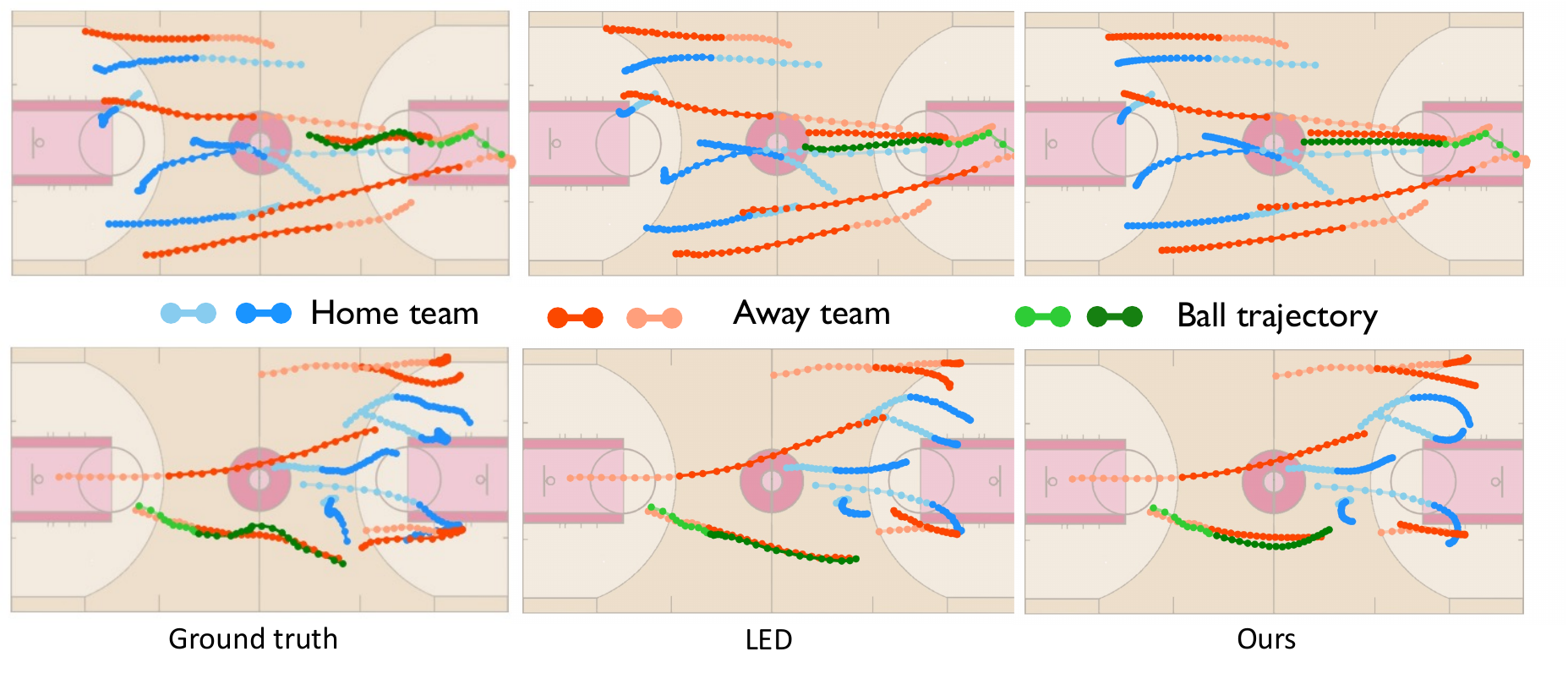}
    \vspace{-0.8cm}
    \caption{
    More qualitative results on the NBA dataset show a comparison between the best-of-20 predictions from our \model IMLE distillation method, the best-of-20 predictions the LED method, and the ground truth future trajectories. The visualization demonstrates that our approach produces predictions that more closely align with the ground truth trajectories compared to the LED model. (Light color indicates past trajectory while dark color means future trajectory; blue/orange/green color: two teams and the basketball)
    }
    \label{fig:qualresult_accuracy_nba_more}
\end{figure*}

\begin{table}[h]
\centering
\caption{Ablations on IMLE sample size $m$ on the \textbf{NBA} dataset. 
We report min\textsubscript{20}ADE/min\textsubscript{20}FDE (meters) for performance.}
\label{tab:ablation_imle_m}
\resizebox{\linewidth}{!}{
\begin{tabular}{c|cccc}
\toprule
\multirow{1}{*}{Trajectory Time} & $m=5$ & $m=10$ & $m=20$ & $m=40$\\
\midrule
1.0s & {0.22}/{0.27} & {0.19}/{0.26} & \textbf{0.18}/\textbf{0.25} & OOM \\
2.0s & {0.38}/\textbf{{0.42}} & \textbf{0.35}/{0.48} & \textbf{0.35}/0.47 & OOM \\
3.0s & {0.58}/{0.70} & {0.55}/{0.69} & \textbf{0.52}/\textbf{0.67} & OOM \\
Total (4.0s) & {0.78}/{0.95} & {0.73}/0.89 & \textbf{0.71}/\textbf{0.87} & OOM \\
\bottomrule
\end{tabular}
}
\vspace{-0.3cm}
\end{table}

\subsection{Model Training Details}
\label{app:train_details}
We preprocess the trajectories using a simple min-max normalization technique, scaling future relative motion to the range $[-1,1]$ linearly to facilitate the training of the flow-matching model. A basic transformer architecture, depicted in 
Fig 1., serves as the backbone of our approach. We find the standard transformer implementation to be sufficient for our task, requiring no special training tricks to achieve strong performance. To facilitate the learning of inter-agent spatial relationships, we apply sinusoidal positional encoding across agents, assigning unique relative positions to each agent's representation. In the motion decoder, we reinforce the agent-level positional encoding and introduce an additional positional encoding at the prediction level. Self-attention is then applied alternately at the agent and prediction levels, strengthening inter-agent interactions and improving overall scene coherence.
The student model trained with IMLE objective shares the same architecture as the teacher model, except that it does not need the flow time positional encoding layers.
For teacher model sampling, we adopt 100 steps to solve the denoising ODEs and generate samples. These samples are used both to evaluate teacher model performance and to train the IMLE distillation models. To enhance IMLE training efficiency, we compute and save the teacher model samples in advance.
All the training is conducted on NVIDIA RTX6000 and A40 GPUs using the AdamW~\cite{loshchilov2018decoupled} optimizer in PyTorch~\cite{paszke2019pytorch}, with weight decay set to 0.01. 

\begin{figure*}[h]
\centering
\includegraphics[width=\linewidth]{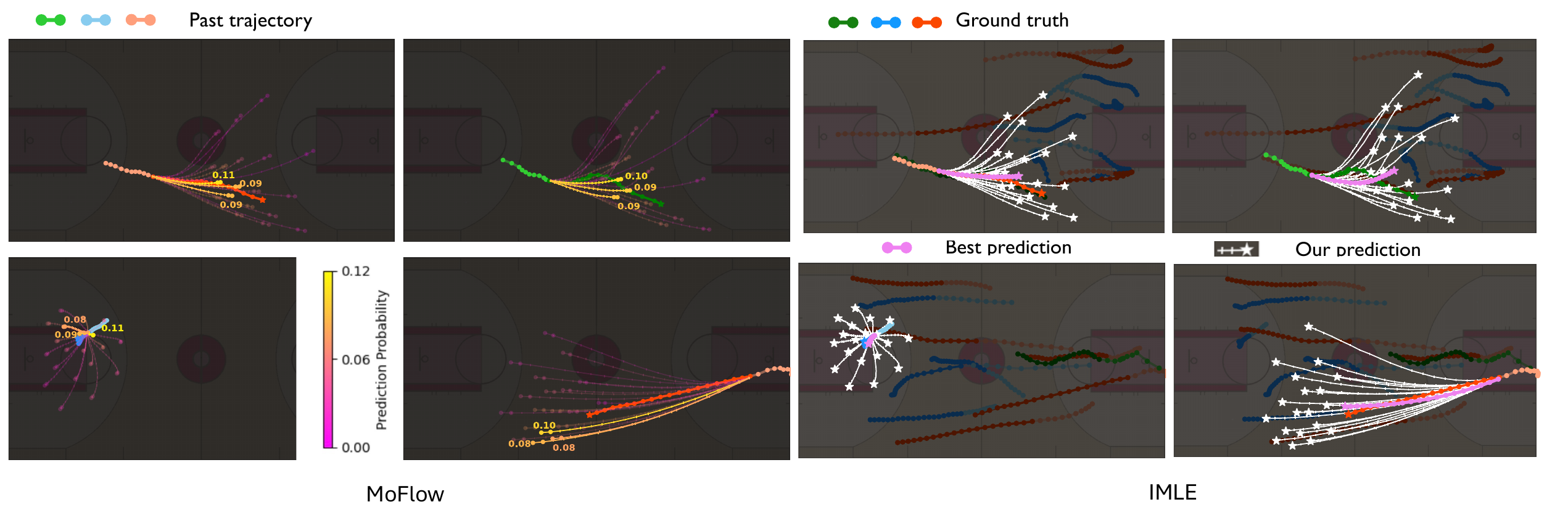}
\caption{Some scenes from NBA dataset with MoFlow predictions for one agent and corresponding IMLE predictions.}
\label{fig:qual2}

\end{figure*}
\begin{figure*}[t]
\centering
\includegraphics[width=\linewidth]{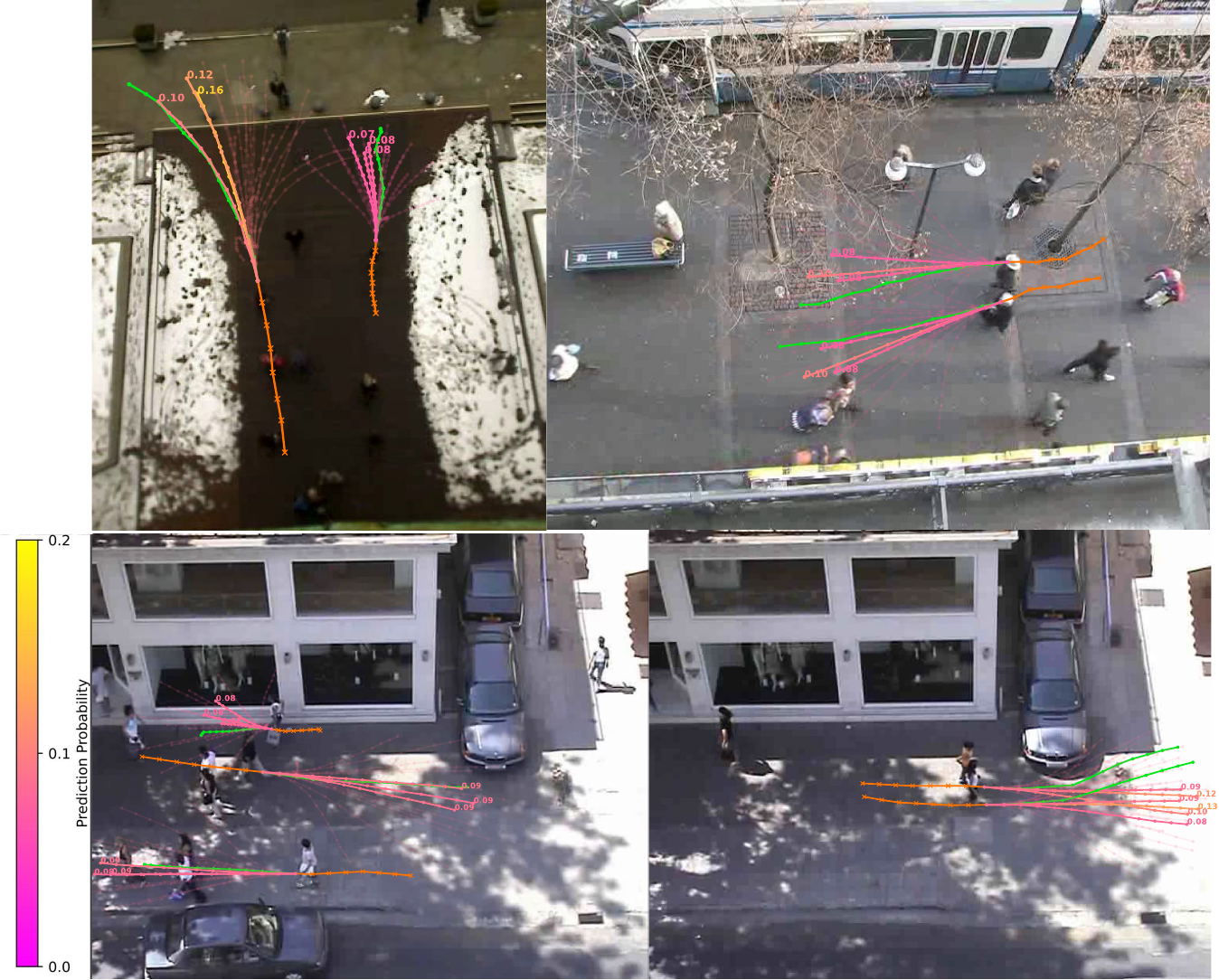}
\caption{Scenes from ETH-UCY datasets with MoFlow generation. The probability besides the trajectories are normalized from classification logit $\{\zeta_i\}_{i=1}^K$ via Softmax. }
\label{fig:qual1}
\end{figure*}

\begin{figure}
  \centering
    \vspace{-0.5cm}
    \includegraphics[width=\linewidth]{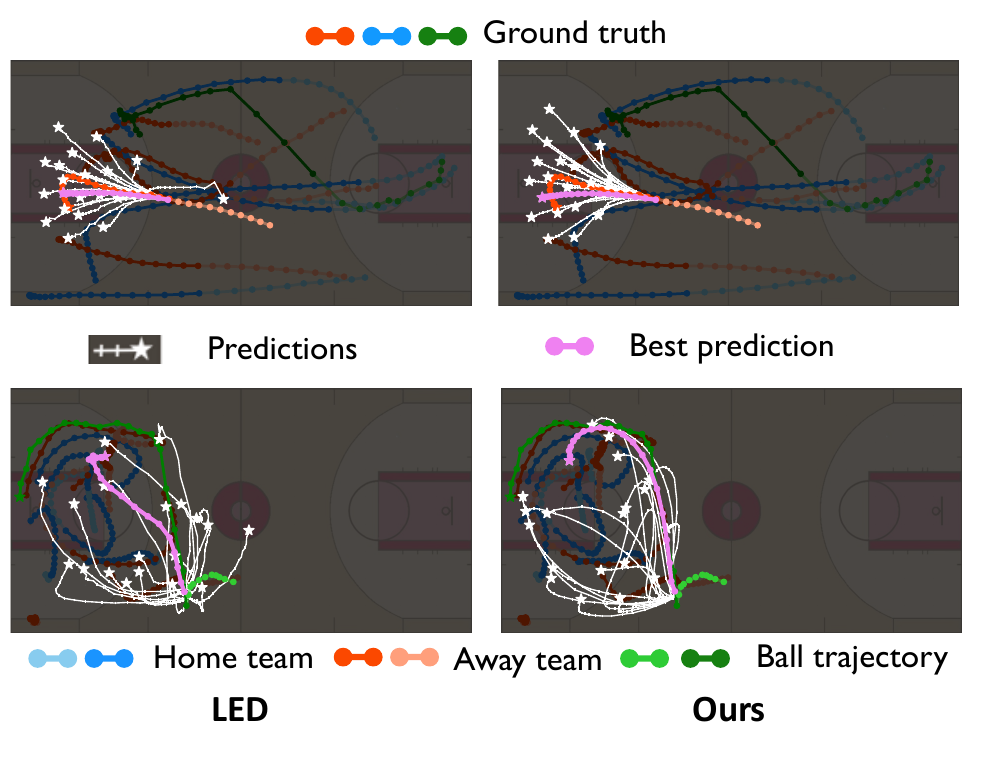}
    \vspace{-0.75cm}
    \caption{More qualitative results on NBA dataset in terms of diversity. We compare our model with LED again.}
  \label{fig:supp_led_nba_diversity}
\end{figure}

\begin{table*}[h]
\centering
\caption{ 
Comparison with baseline distillation models on \textbf{ETH-UCY} dataset. min\textsubscript{20}ADE/min\textsubscript{20}FDE (meters) are reported. Bold/underlined fonts represent the best/second-best result. Note that we evaluate our approach on this dataset with the same split as SGAN~\cite{gupta2018social} while~\cref{tab:eth_ucy_result} in the main text is based on the dataset and train-test split released by the LED~\cite{Mao_2023_CVPR} group.}
\label{tab:eth-ucy-original-dataset}
\vspace{-2pt}
\resizebox{\linewidth}{!}{%
\begin{tabular}{c|ccccccc|cc}
\toprule
\multirow{3}{*}{Subsets} &MID&GroupNet& TUTR& EqMotion& EigenTraj & LED&SingularTraj&\textbf{\model} & \textbf{IMLE}   \\
&~\cite{Gu_2022_CVPR} &~\cite{groupnet_CVPR} &~\cite{Shi_2023_ICCV} &~\cite{xu2023eqmotion} &~\cite{bae2023eigentrajectory} & ~\cite{Mao_2023_CVPR}&~\cite{bae2024singulartrajectory} & \\
\midrule
ETH &  0.39/0.66 & 0.46/0.73 & 0.40/0.61 &0.40/0.61 &\u{0.36}/\u{0.53} & {0.39}/0.58 & \textbf{0.35}/\textbf{0.42} & 0.39/0.55 & 0.40/0.61 \\ 
HOTEL &   0.13/0.22 & 0.15/0.25 & \textbf{0.11}/\u{0.18} & \u{0.12}/\u{0.18}&\u{0.12}/0.19 & \textbf{0.11}/\textbf{0.17} &0.13/0.19 & \textbf{0.11}/\textbf{0.17} & 0.12/0.18 \\ 
UNIV &  \textbf{0.22}/0.45 & 0.26/0.49 & \u{0.23}/0.42 & \u{0.23}/0.43&0.24/0.34 & 0.26/0.44 & 0.25/0.44 & \u{0.22}/\textbf{0.39} & 0.24/0.44 \\ 
ZARA1 &  0.17/0.30 & 0.21/0.39 & 0.18/0.34 &0.18/0.32 & 0.19/0.33 & 0.18/\textbf{0.26} & 0.19/0.32 & \textbf{0.17}/\u{0.29} & \textbf{0.17}/0.31\\ 
ZARA2 &  \u{0.13}/0.27 & 0.17/0.33 & \u{0.13}/0.25 &\u{0.13}/\u{0.23}& {0.14}/0.24 & \u{0.13}/\textbf{0.22} &0.15/0.25 & \textbf{0.12}/\textbf{0.22} & 0.13/0.24 \\ 
\midrule
AVG  &  \u{0.21}/0.38 & {0.25}/0.44 & \u{0.21}/0.36 &\u{0.21}/0.35 & \u{0.21}/\u{0.33} & \u{0.21}/\u{0.33} & \u{0.21}/\textbf{0.32} & \textbf{0.21}/\textbf{0.32} & 0.21/0.36\\ 
\bottomrule
\end{tabular}
}
\end{table*}

\begin{table}[b]
\vspace{-0.5cm}
\centering
\caption{Sampling speed-up comparison on the \textbf{NBA} dataset. We report min\textsubscript{20}ADE/min\textsubscript{20}FDE (meters) for empirical performance. DDPM refers to the teacher diffusion model used in LED~\cite{Mao_2023_CVPR}.}
\label{tab:ablation_nfe}
\resizebox{\linewidth}{!}{
\begin{tabular}{c|cc|ccc}
\toprule
\textit{Efficiency} & \textbf{\model} & \textbf{IMLE} & \textbf{DDPM} & \textbf{LED} \\
\midrule
NFE & {100} & 1 & {100} & 6 \\
Runtime (ms) & 33.20 & 0.70 & 796 & 22.38 \\
\midrule
\textit{Performance} & \textbf{\model} & \textbf{IMLE} & \textbf{DDPM} & \textbf{LED}\\
\midrule
1.0s & {0.18}/{0.25} & {0.18}/{0.25} & {0.20}/{0.28} & {0.18}/{0.27} \\
2.0s & {0.34}/{0.47} & {0.35}/{0.47} & {0.43}/{0.64} & {0.37}/{0.56} \\
3.0s & {0.52}/{0.67} & {0.52}/{0.67} & {0.68}/{0.95} & {0.58}/{0.84} \\
Total (4.0s) & {0.71}/{0.87} & {0.71}/0.87 & {0.93}/{1.20} & {0.81}/{1.10} \\
\bottomrule
\end{tabular}
}
\end{table}

\section{Further Qualitative Results}
\label{app:qual}
\begin{figure*}[t]
  \centering
    \vspace{-0.75cm}
    \includegraphics[width=\linewidth]{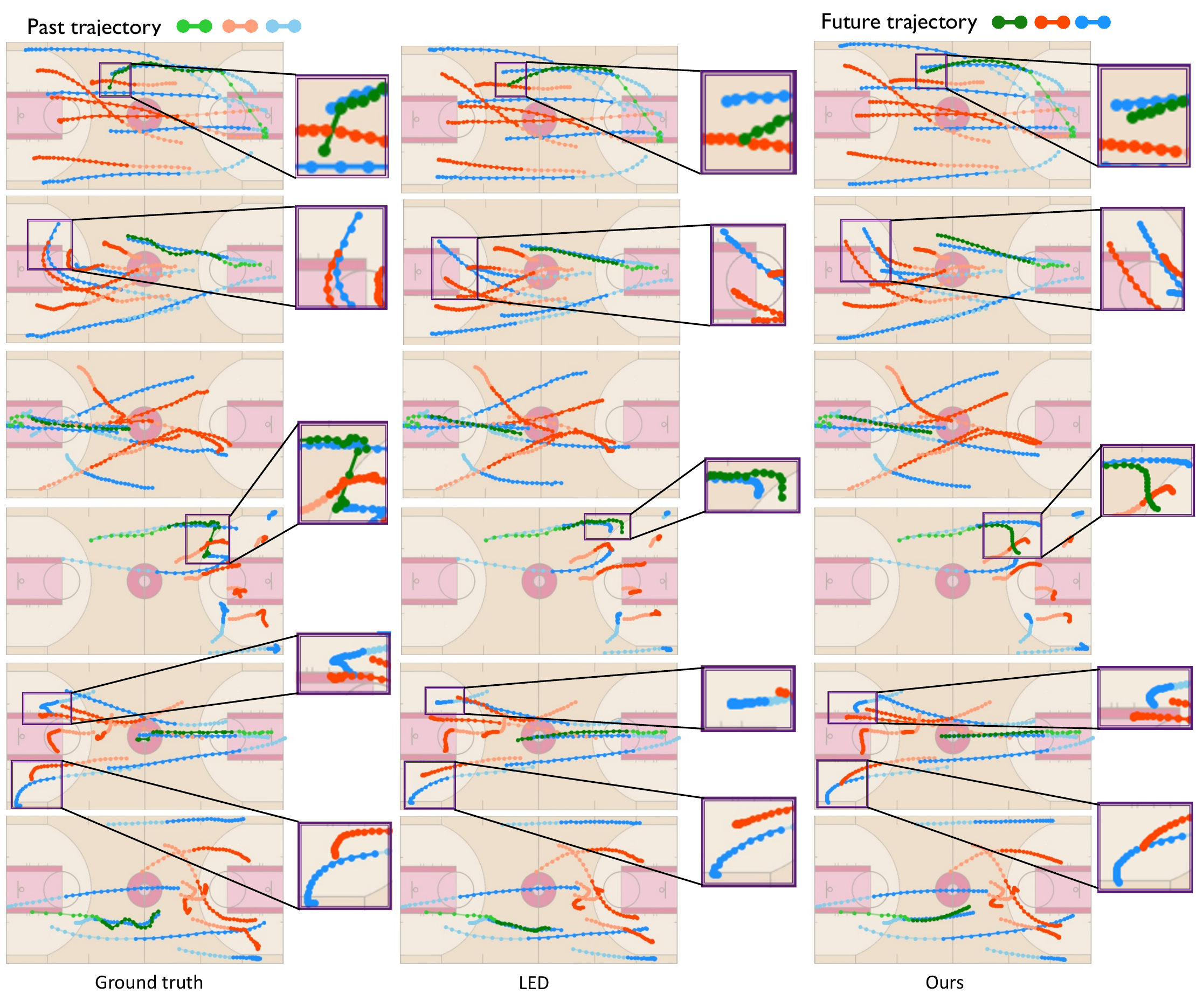}
    \vspace{-0.75cm}
    \caption{More qualitative results on the NBA dataset show a comparison among the best-of-20 predictions from our \model IMLE distillation method, the best-of-20 predictions the LED method, and the ground truth future trajectories. The visualization demonstrates that our approach produces predictions that more closely align with the ground truth trajectories compared to the LED model. The key differences are zoomed in from the purple rectangles and displayed side-by-side, contrasting the ground truth, the best samples generated by the LED model with those produced by our approach. (blue/orange/green color: home team, away team and the basketball)}
  \label{fig:supp_led_nba_accuracy}
\end{figure*}

We would love to show more qualitative results on NBA dataset. In~\cref{fig:supp_led_nba_20_compare}, we compare our 20 predictions with the samples\footnote{Note that these are the exactly two same scenes with identical ego agents, as demonstrated by their paper and \href{https://github.com/MediaBrain-SJTU/LED/blob/main/visualization/draw_mean_variance.ipynb}{codebase}. The coloring has been adjusted to align with our visualizations, and we have replaced their mean prediction with the best-of-20 prediction for consistency.} generated by LED~\cite{Mao_2023_CVPR} model. Upon closer examination of the predictions generated by the LED model, we observe that certain trajectories, which highlighted by the red circles, move in implausible, opposite directions. In contrast, our model effectively captures the general movement, steering the trajectories towards more plausible directions. Notably, the two figures in the first row of the LED model predictions are classified as backcourt violations defined in the basketball rules. Such violations are rare in the NBA training dataset. Notice that violations are not present in our predictions, indicating that our model generates trajectories that are more realistic and contextually appropriate for basketball games. 
According to~\cref{fig:supp_led_nba_accuracy}, we have successfully predicted the direction of the future trajectories given the context on a regular basis while other state-of-the-art method failed to achieve. In particular, we highlight the subtle yet essential differences among the best-of-20 predictions from our MoFlow IMLE distillation method, the best-of-20 predictions from the LED method, and the ground truth future trajectories by zooming in on the details. 

It is important to highlight that our \model model can generate $K$ scene-level trajectory predictions, with corresponding classification logits \(\{\zeta_i\}_{i=1}^K, \zeta_i \in \mathbb{R}\), depicted by~\cref{fig:qual1} and~\cref{fig:qual2}. This means we have an empirical distribution for all the trajectories that we generate for each agent. 
While the predicted trajectories may sometimes be widely distributed, our \model model effectively differentiates their plausibility. 
\section{Additional Experimental Results}
\label{app:additional_exp}
In this section, we conduct ablation experiments on the NBA dataset to validate our model design choices. Due to a recent paper~\cite{weng2023joint}, we will report another set of results under the joint metric to show the capacity of our models.


\subsection{Sampling Speed-up Effect}
\label{app:speed_up}
We demonstrate the sampling speed-up effect in~\cref{tab:ablation_nfe}, reporting the average sampling time per scene for runtime. Thanks to the one-step inference enabled by IMLE distillation, our student model achieves significantly faster sampling with 100x fewer NFEs, reducing runtime by 98\% on the same hardware. Notably, we found that IMLE distillation does not compromise empirical performance in our setup. Moreover, both our teacher and student model achieve faster sampling speeds than state-of-the-art methods, while delivering superior performance.

\subsection{Ablations on Flow Matching Configurations}
To assess the significance of each module, we conduct an ablation study on different components of our \model. In 
Section 4.2, we discussed the design of input-output dimension adaptation and the time schedule. Here, we demonstrate the superiority of our current configuration through the results presented in~\cref{tab:ablation_moflow}.

From~\cref{tab:ablation_moflow}, we observe that positional encoding applied at the predictions ($K$) level provides a greater advantage compared to positional encoding applied at applied at the agent level. Additionally, incorporating shared noise yields strong final results without leading to the variance explosion observed in the IID column.

\subsection{Ablations on IMLE Configurations}
\label{app:imle_ablations}

According to the IMLE training principle outlined in 
Algorithm 1,
the student model needs to sample $m > 1$ instances to select the one closest to the teacher model's sample. We present the comparison results for different values of $m$ in~\cref{tab:ablation_imle_m}. We choose $m=20$ for its superior empirical performance.
Note that $m$ effectively enlarges the training-time mini-batch size by a factor of $m$. Therefore, setting $m$ too large can lead to out-of-memory issues. Moreover, we present the JADE/JFDE results and conduct ablation on $\omega$ in by re-training our MoFlow with the new objective, which is identical to equations (10,11)~\cite{weng2023joint}. Based on ~\cref{tab:joint_metric}, we observe a trade-off between these two metrics: a marginal improvement in JADE/JFDE leads to a significant drop in ADE/FDE performance. Next, we assess our model's ability to preserve the sample quality of the teacher model. Specifically, when analyzing the metric map-aware self-distance (MASD), we observe that the LED model fails to maintain this quantity, exhibiting a substantial deviation of $\sim$9m from the teacher model (DDPM). In contrast, our IMLE model preserves this quantity with remarkable accuracy, achieving a deviation of only $\sim$0.1m. These results further underscore the superiority of our approach in the distillation task.

\end{document}